%% file: main.tex
\documentclass[twocolumn]{IEEEtran}

\usepackage[pdftex]{graphicx} 
\usepackage{amsmath} 
\usepackage{algorithmic}
\usepackage{array} 
\usepackage[caption=false,font=footnotesize]{subfig} 
\usepackage{dblfloatfix}
\usepackage{url} 
\usepackage{microtype}
\usepackage[utf8]{inputenc} 
\usepackage[T1]{fontenc}    
\usepackage{paralist} 
\usepackage{marvosym} 
\usepackage{bm} 
\usepackage{bbm} 
\usepackage{amssymb}   
\usepackage{mathrsfs} 
\usepackage{tabularray} 
\UseTblrLibrary{booktabs}       
\usepackage{tabularx} 
\usepackage[flushleft]{threeparttable} 
\usepackage{colortbl} 
\usepackage{makecell} 
\usepackage{multirow} 
\usepackage{nicefrac} 
\usepackage[dvipsnames]{xcolor} 
\usepackage{soul} 
\setul{0.5ex}{0.3ex}

\definecolor{citecolor}{RGB}{34,139,34}
\usepackage[pagebackref=true,breaklinks=true,letterpaper=true,colorlinks,
    citecolor=citecolor,bookmarks=false]{hyperref}
\usepackage{cleveref}  
\crefformat{figure}{Fig.~#2#1#3}
\crefformat{equation}{Eq.~(#2#1#3)}
\crefformat{table}{Table~#2#1#3}
\crefformat{section}{Section~#2#1#3}
\crefformat{algorithm}{Algorithm~#2#1#3}
\crefformat{appendix}{Appendix #2#1#3}
\crefformat{subsection}{subsection~#2#1#3}
\usepackage{cellspace}
\usepackage[numbers,sort&compress]{natbib} 

\usepackage{xspace} 

\usepackage{pifont} 
\definecolor{modifiedorange}{RGB}{255,140,0} 

\usepackage{float}
\usepackage{placeins}

\definecolor{Gray}{rgb}{0.9,0.9,0.9}
\definecolor{LightCyan}{rgb}{0.88,1,1}
\newcolumntype{a}{>{\columncolor{Gray}}c}
\newcolumntype{b}{>{\columncolor{white}}c}

\begin{document}

\setlength{\abovedisplayskip}{.5\baselineskip} 
\setlength{\belowdisplayskip}{.5\baselineskip} 

\title{
SeqCSIST: Sequential Closely-Spaced Infrared\\ Small Target Unmixing
}

\input{contents/1-authors.tex}

\maketitle

\input{contents/2-abstract.tex}
\input{contents/3-introduction.tex}

\input{contents/4-related-work.tex}

\input{contents/5-dataset.tex}

\input{contents/6-method.tex}
\input{contents/7-experiment.tex}
\input{contents/8-analysis.tex}
\input{contents/9-conclusion.tex}

\bibliographystyle{IEEEtran}
\bibliography{./reference.bib}

\input{contents/10-biography.tex} 

\end{document}

%% file: contents/1-authors.tex

\author{
  Ximeng~Zhai,
  Bohan~Xu,
  Yaohong~Chen,
  Hao~Wang,
  Kehua~Guo,
  Yimian~Dai
  \thanks{
    This work was supported by
      the National Natural Science Foundation of China (62301261, 
      62472443) 
      and the West Light Foundation of the Chinese Academy of Sciences (XAB2022YN06).
    \emph{
    (Corresponding author: Yaohong Chen and Yimian Dai).}
    }

  \thanks{
    Ximeng Zhai, Yaohong Chen and Hao Wang are with Xi'an Institute of Optics and Precision Mechanics, Chinese Academy of Sciences, Xi'an, China.
    (e-mail:
    \href{mailto:zhaiximeng23@mails.ucas.ac.cn}{zhaiximeng23@mails.ucas.ac.cn};
    \href{mailto:chenyaohong@opt.ac.cn}{chenyaohong@opt.ac.cn};
    \href{mailto:wanghao@opt.ac.cn}{wanghao@opt.ac.cn}).
  }
  
  \thanks{Bohan Xu is with College of Information Science and Engineering, Henan University of Technology, Zhengzhou, China.
  }

  

  \thanks{
    Kehua Guo is with School of Computer Science and Engineering, Central South University, Changsha 410083, China.
    (e-mail: \href{mailto:guokehua@csu.edu.cn}{guokehua@csu.edu.cn}).
  }
  
  \thanks{
  Yimian Dai is with VCIP, College of Computer Science, Nankai University. He also holds a position at the NKIARI, Shenzhen Futian. (e-mail:
  \href{mailto:yimian.dai@gmail.com}{yimian.dai@gmail.com}).
  }
  
  
}

%% file: contents/2-abstract.tex
\begin{abstract}
Due to the limitation of the optical lens focal length and the resolution of the infrared detector, distant Closely-Spaced Infrared Small Target (CSIST) groups typically appear as mixing spots in the infrared image.
In this paper, we propose a novel task, 
Sequential CSIST Unmixing,
namely detecting all targets in the form of sub-pixel localization from a highly dense CSIST group.
However, achieving such precise detection is an extremely difficult challenge.
In addition, the lack of high-quality public datasets has also restricted the research progress. 
To this end, firstly, we contribute an open-source ecosystem, including SeqCSIST, a sequential benchmark dataset, and a toolkit that provides objective evaluation metrics for this special task, along with the implementation of $23$ relevant methods.
Furthermore, we propose the Deformable Refinement Network (DeRefNet), a model-driven deep learning framework that introduces a Temporal Deformable Feature Alignment (TDFA) module enabling adaptive inter-frame information aggregation.
To the best of our knowledge, this work is the first endeavor to address the CSIST Unmixing task within a multi-frame paradigm.
Experiments on the SeqCSIST dataset demonstrate that our method outperforms the state-of-the-art approaches with mean Average Precision (mAP) metric improved by 5.3\%. Our dataset and toolkit are available from \url{https://github.com/GrokCV/SeqCSIST}.
\end{abstract}


\begin{figure}[h]
    \flushright 
    \includegraphics[width=.48\textwidth]{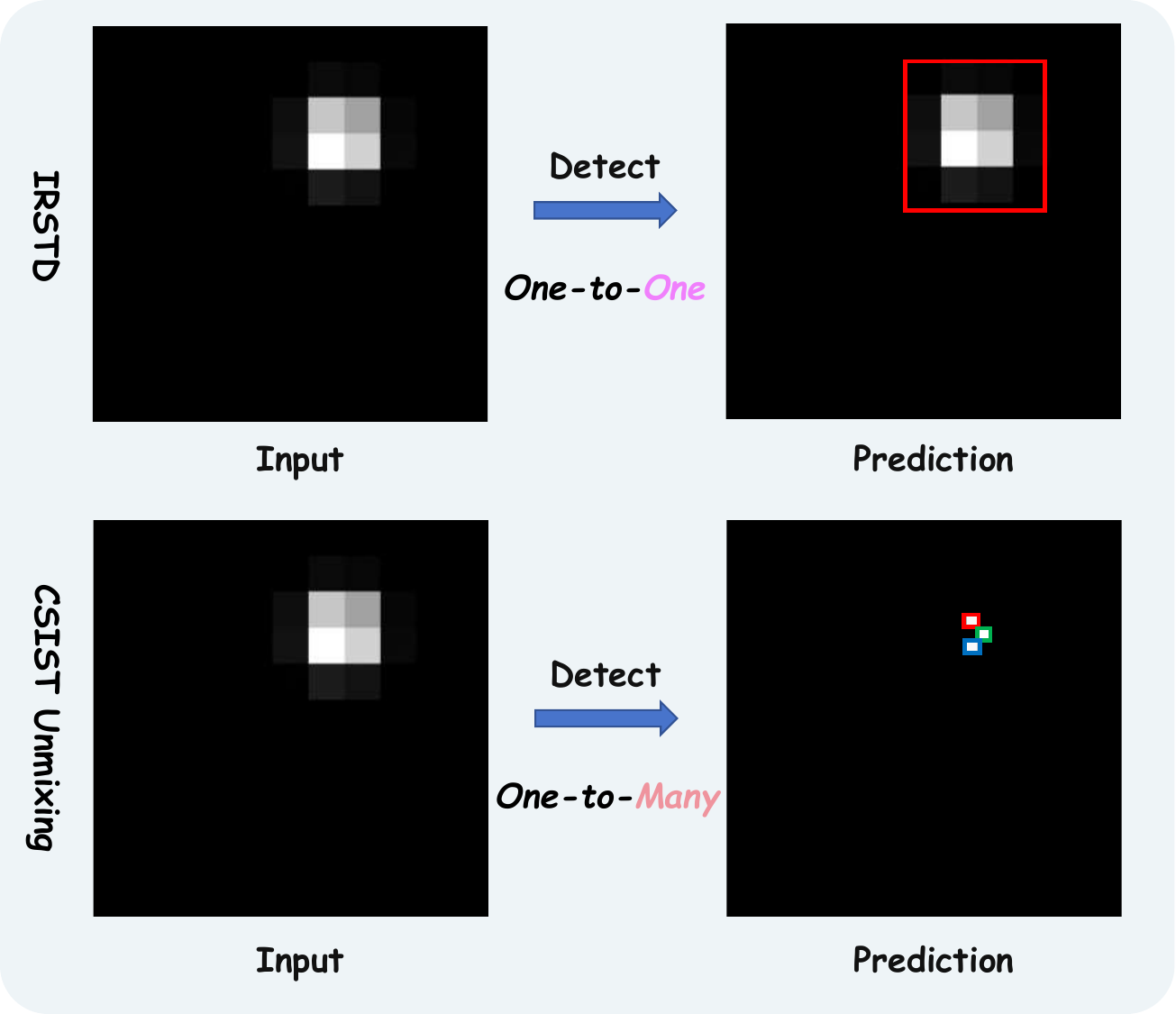}
    \caption{The first row illustrates that IRSTD methods assume a one-to-one correspondence between detected targets and real-world objects. When there is only one spot in the image to be recognized, the detection result corresponds to a single target. In contrast, the second row for CSIST Unmixing performs more refined detection, enabling sub-pixel localization and unmixing of potential sub-targets within the spot.}
    \label{fig:IRSTD_VS_CSISTU}
\end{figure}

\begin{figure*}[htbp]
    \centering
    \includegraphics[width=.98\textwidth]{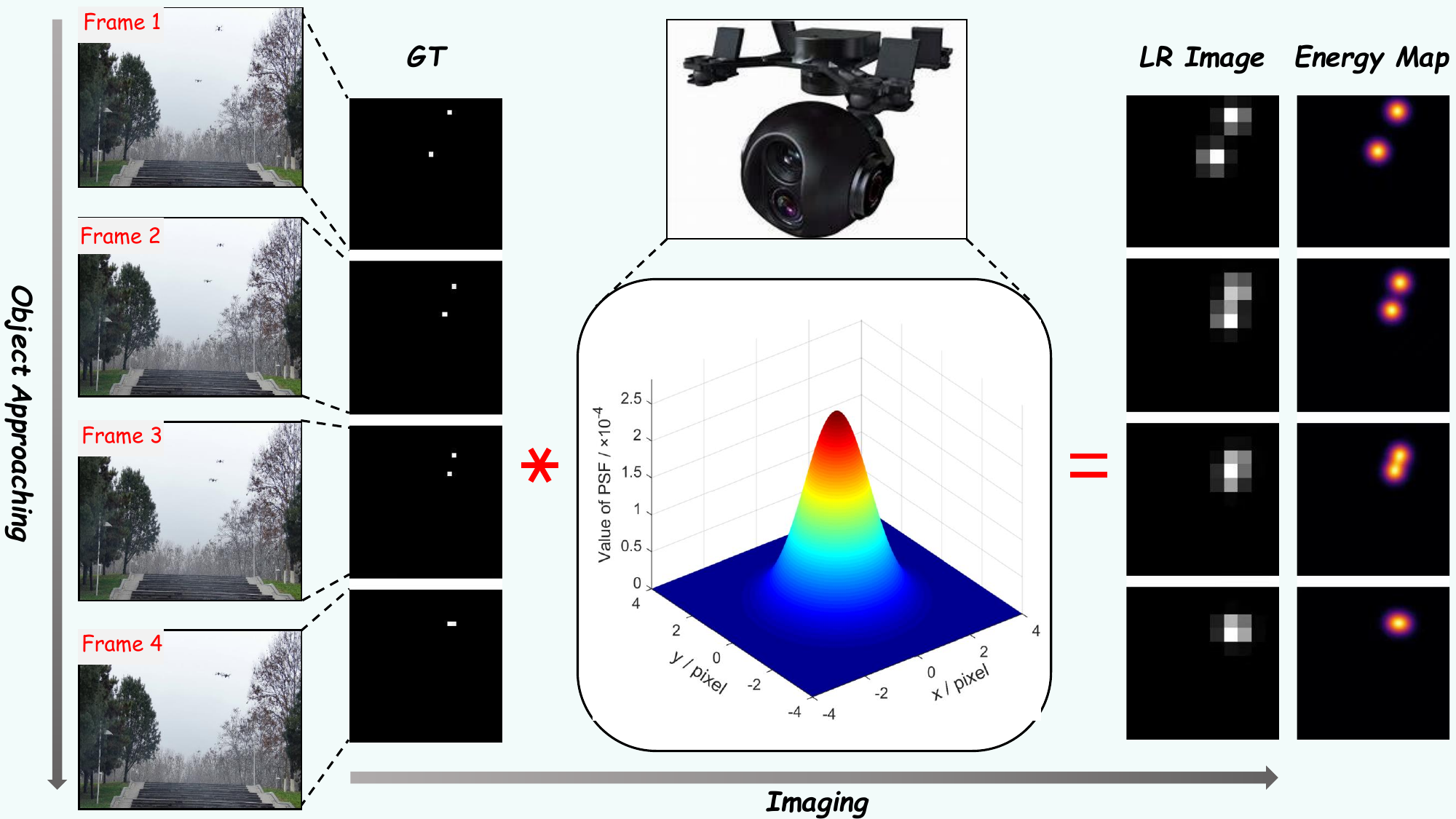}
    \caption{The diagram shows that as the long-distance small targets get closer, the energy mixing becomes more and more obvious, and it is no longer possible to distinguish the number and specific location of the targets through visual observation}
    \label{fig:CSIST_PSF}
\end{figure*}

%% file: contents/3-introduction.tex

\section{Introduction} \label{sec:introduction}


\IEEEPARstart{I}{nfrared} imaging remains unaffected by lighting variations, making it highly reliable for capturing key scenes and objects even in challenging low-light conditions. This unique capability ensures consistent performance across diverse environments \cite{wang2024improving}.
Thus, infrared imaging plays a vital role in both military and civilian applications, such as surveillance, border protection, and search-and-rescue operations. 
However, due to the requirements of early warning detection systems and the resolution limitations of infrared imaging systems, long-distance targets of interest often appear as small  objects within the field of view.
Consequently, the ability to precisely detect infrared small targets from a distance is crucial for the success of these applications \cite{dai2024pick}.

Typically, Infrared Small Target Detection (IRSTD) focuses on searching and detecting targets in large-scale areas, where several critical limitations emerge:
\begin{enumerate}
    \item 
    \textbf{Lack of Intrinsic Feature:} 
    Owing to the substantial distance between the sensor and the target, the latter typically presents itself as a low-contrast entity \cite{zhang2022isnet}. The absence of distinct features, such as texture or shape, further complicates the detection process.
    \item \textbf{Overlap of Dense Targets:} Densely packed targets in infrared images can cause overlapping signatures, making it difficult for detection algorithms to distinguish and identify individual targets.
\end{enumerate}

To solve these challenges, deep learning approaches \cite{dai2021attentional} harness end-to-end feature extraction in high-dimensional spaces, systematically improving detection accuracy and operational robustness. This has shifted the focus towards multi-scale feature fusion, bridging the gap between detailed target representation and contextual information.
Earlier effort, such as Dense Nested Attention Network (DNANet) \cite{li2022dense}, achieves multi-scale feature extraction and meticulous detail preservation by leveraging dense cross-layer semantic interactions. 
Subsequently, whether through multi-scale progressive fusion in semantic segmentation, as exemplified by U-Net in U-Net (UIU-Net) framework \cite{wu2022uiu}, or the regression-based detection paradigm, as seen in Local-to-Global fusion network (LoGoNet) \cite{li2023logonet}, both approaches enhance detection by leveraging multi-scale fusion with task-specific structural optimizations.

While the aforementioned IRSTD methods have achieved certain progress and advanced the field, their applications are primarily \textbf{limited to scenes of large field-of-view but sparse targets}.
The assumption behind these conventional IRSTD methods is that there is \textit{an exact one-to-one mapping between the detected objects in the image and their corresponding real-world counterparts}, as shown in the first row of Fig.~\ref{fig:IRSTD_VS_CSISTU}.

Due to the sensor resolution constraint, as shown in Fig.~\ref{fig:CSIST_PSF}, \textit{when the spatial distance of targets gets closer, their energy gradually overlaps during infrared imaging}, \textbf{inevitably stacking the neighboring targets onto the same single pixel} and resulting in blurred spots on the imaging plane \cite{lin2011resolution}.
In such scenarios, above methods fail to deliver the desired detection performance. Additionally, previous detection methods focus on identifying target contours, which are often absent in point targets. Even if spot contours of dense targets are detected, the exact number and positions of targets cannot be determined.

To fill the critical gaps in this field, we propose \textbf{a novel task}:  Closely-Spaced Infrared Small Targe Unmixing. 
It is crucial to recognize that \textit{this task differs fundamentally from traditional IRSTD} in its nature and scope, which is \textbf{a downstream task of IRSTD}.
While IRSTD provides a framework for small target detection, it struggles with complex scenes involving high-density targets. 
CSIST Unmixing takes this challenge further by explicitly addressing such scenarios, paving the way for more accurate and reliable detection in real-world applications. 
Unlike conventional detection tasks focus on target context and features, this new task is characterized by the following:
\begin{enumerate}
    \item \textbf{Handling of Dense Target Scenarios:}  IRSTD focuses on detecting sparse, isolated targets in infrared imagery. In contrast, CSIST Unmixing handles scenarios with closely packed targets and energy aliasing, focusing on \textit{resolving target groups that are visually indistinguishable}. 
    \item \textbf{Sub-Pixel Detection Precision:} As shown in the Fig.~\ref{fig:IRSTD_VS_CSISTU}, 
    unlike IRSTD that detects targets in form of bounding boxes,
    CSIST Unmixing decomposes blurred target spots in the pixel space into distinct sub-targets, \textit{transforming the task from simple detection to precise sub-pixel localization}.
\end{enumerate}
This makes CSIST Unmixing a significantly more challenging task compared to conventional IRSTD.
Naturally, a question arises: \textit{Can we achieve better CSIST Unmixing results through spatiotemporal information fusion?}

Our answer is affirmative. 
In this paper, we first construct a sequence-based benchmark dataset, SeqCSIST, which serves as the foundation of our open-source ecosystem.
The SeqCSIST dataset encompasses 100,000 frames organized into 5,000 random trajectories. 
Each trajectory comprises 20 consecutive target crops with a size of $11 \times 11$ pixels, where overlapped target number varies from $2$ to $4$.
Noted that SeqCSIST is the first dataset specially for Sequential CSIST Unmixing research.
Besides the dataset, we also release a toolkit as the second cornerstone of our open-source ecosystem.
This toolkit includes a metric for objectively evaluating capabilities of CSIST Unmixing, along with the implementation of $23$ relevant methods.


To address the inherent challenges of this novel task, we propose the \textbf{De}formable \textbf{Ref}inement \textbf{Net}work (\textbf{DeRefNet}) to achieve effective Sequential CSIST Unmixing, which consists of three main modules: a sparsity-driven feature extraction module, a positional encoding module and a Temporal Deformable Feature Alignment (TDFA) module.
Distinct from conventional approaches that rely on generic ResNet backbones for feature extraction, DeRefNet fully considers the sparsity prior of targets and achieves effective extraction of CSIST features through nonlinear learnable and sparsifying transforms.
Subsequently, to enable finer sub-pixel target localization, a positional encoding module is utilized to enhance temporal information. 
Finally, the TDFA module
enables dynamic reference-based refinement for middle frame, which is processed through multi-frame deformable alignment at a feature level without explicit motion estimation and image wrapping operations.
Noted that DeRefNet marks the pioneering attempt to incorporate a multi-frame approach into the field of CSIST Unmixing.

The main contributions of this paper can be summarized as follows:
\begin{enumerate}
    \item \textbf{Novel Task:} 
    We propose a novel task, Sequential CSIST Unmixing, which broadens the definition of IRSTD.
    \item \textbf{Open-Source Ecosystem:} 
    We release an open-source ecosystem for CSIST Unmixing, including the SeqCSIST dataset and a toolkit for benchmarking.
    \item \textbf{End-to-End Framework:} 
    We propose DeRefNet, a multi-frame model-driven deep learning architecture with temporal deformable feature alignment.
\end{enumerate}

%% file: contents/4-related-work.tex
\section{Related Work} \label{sec:related}

\subsection{Sequential Infrared Small Target Detection}

Over the past few years, Sequential Infrared Small Target Detection (SIRST Detection) has emerged as a crucial technology in infrared warning and tracking systems. It is valued for its high sensitivity and resistance to interference \cite{tong2024st}, enabling the detection and tracking of small targets in continuous frames.

Traditional SIRST Detection approaches, such as local-contrast-based \cite{chen2013local}, human visual-based \cite{han2014robust}, and low-rank-based \cite{feng2023coarse} methods, have shown effectiveness in specific conditions but struggle with common challenges like complex scenes. 
To address these limitations, researchers have developed deep learning-based methods which enhance detection accuracy and robustness.
Recent works, Chen \emph{et al.} proposed the Memory Enhanced Global-Local Aggregation (MEGA) network \cite{chen2020memory}. It represents the first method to systematically integrate both global contextual information and local detail features in infrared target detection.  
While this approach enhances keyframe feature representation, it relies heavily on empirically designed structures. Similarly, Zhang \emph{et al.}'s Time-Aware Fully Convolutional neural network (TFCST) \cite{huang2024lmaformer} and Deng \emph{et al.}'s Background Estimation framework (BEmST) \cite{deng2024bemst} advance spatiotemporal modeling but maintain conventional feature extraction paradigms.

Although these approaches show effectiveness in modeling spatiotemporal relationships, they share a common limitation: they primarily focus on designing sophisticated detection heads while adopting general network backbones. Such design overlooks the inherent sparsity characteristics of CSIST, leading to suboptimal feature representations. 
Furthermore, these methods ignore the
domain-specific priors which could potentially guide the learning process towards more effective feature extraction.
Our work addresses these limitations through the following two highlights:
\begin{enumerate}
    \item \textbf{Downstream Task:} 
    Sequential CSIST Unmixing and SIRST Detection are fundamentally different tasks, with CSIST Unmixing focusing on sub-pixel localization while SIRST Detection on identifying target contours. Our work highlights this distinction by introducing a sequential benchmark for CSIST Unmixing.
    \item \textbf{Model-Driven Backbone Network:} 
    Embedding domain-specific prior knowledge via a model-driven backbone network, our framework better addresses the unique challenges of Sequential CSIST Unmixing compared to generic SIRST Detection frameworks.

\end{enumerate}

\subsection{Deep Unfolding}
The deep unfolding paradigm bridges the strengths of model-based methods and deep neural networks by transforming iterative inference processes into layer-wise structures akin to neural networks. 
This transformation allows model parameters to vary across layers, enabling more flexible architectures that can be efficiently optimized through gradient-based methods \cite{zhang2020deep}.
For example, based on the model-based mothod Alternating Direction Method of Multipliers (ADMM) algorithm, ADMM-Net maps each iteration to a neural network layer and optimizes parameters through discriminative learning \cite{yang2018admm}.

Recent years have witnessed increasing applications of deep unfolding across various computer vision tasks. 
Zhang \emph{et al.} leveraged the strengths of both model-based and learning-based approaches by reformulating the Iterative Shrinkage-Thresholding Algorithm (ISTA) into a deep neural network, ISTA-Net \cite{zhang2018ista}. 
While this method addresses the performance bottlenecks of traditional methods that rely on nonlinear sparsifying transforms, it lacks flexibility when dealing with multi-scene images in practical applications.
Building on this work, You \emph{et al.} proposed the ISTA-unfolding deep network (ISTA-Net++), a flexible and adaptive deep deployment network for compressed sensing \cite{you2021ista}. It achieves the image reconstruction across diverse compressed sensing ratios.

We acknowledge the valuable contributions of these attempts, but they have mainly focused on low-level vision tasks, such as image reconstruction. In contrast, our work shifts the focus to the high-level task of CSIST Unmixing, a domain where such approaches have been rarely explored.
Our primary contributions are underscored in the following two key aspects:
\begin{enumerate}
    \item \textbf{Pioneering Use in New Task:} We pioneer introducing the deep unfolding paradigm to Sequential CSIST Unmixing, opening up new possibilities for future design of optimization-driven neural networks in this domain.
    \item \textbf{Dynamic Scenario Adaptability:} Unlike conventional deep unfolding methods, which rely on fixed sampling locations, our model dynamically adjusts the sampling positions condition on the input image, allowing for better adaptation to multi-frame CSIST scenarios.
\end{enumerate}

%% file: contents/5-dataset.tex
\section{SeqCSIST Dataset} \label{sec:dataset}

\begin{figure}[h]
    \flushleft 
    \includegraphics[width=.48\textwidth]{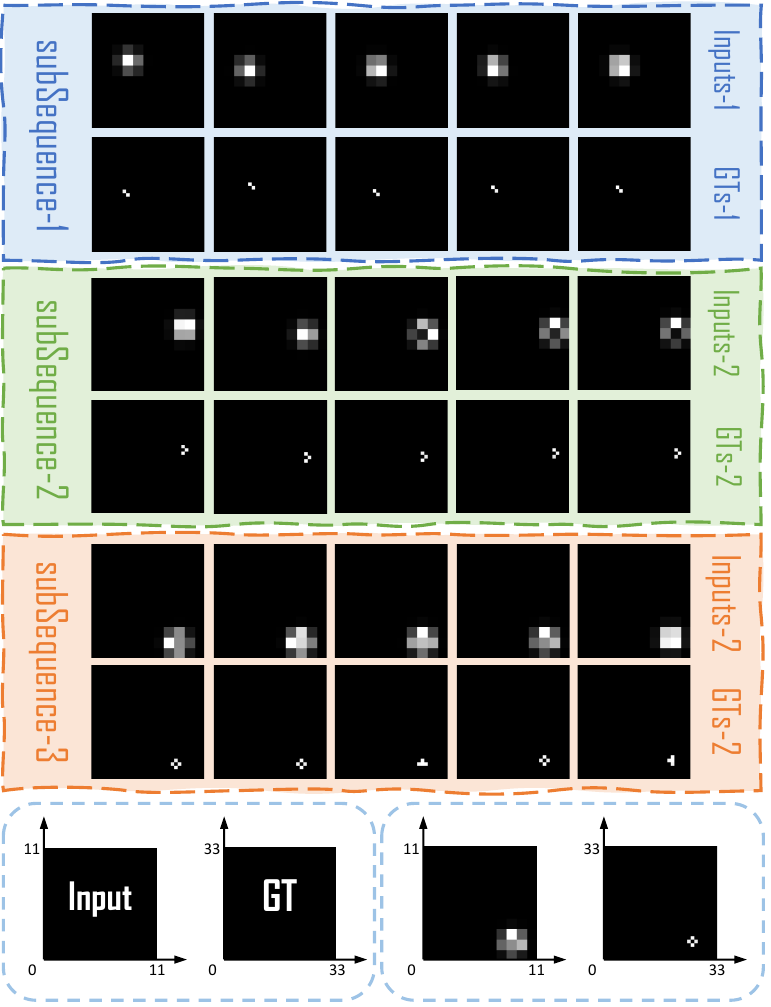}
    \caption{
    The SeqCSIST dataset comprises two components: the first being the $11 \times 11$ low-resolution image captured by the optical system, and the second being the specific coordinate label of the GT. Each row represents a sequence of sequential partial frames, i.e., subsequence. In each subsequence, the first row presents $11 \times 11$ low-resolution images, with their corresponding GT visualizations displayed in the second row at an upsampled resolution of $33 \times 33$ for better visibility of target locations.
    }
    \label{fig:dataset}
\end{figure}

Due to the constraints imposed by the optical lens's focal length and the resolution of the infrared detector, distant targets typically manifest as low-intensity regions and span only a few pixels \cite{liu2024infrared} on the infrared imaging plane.
During imaging process, optical diffraction distributes a point target’s energy to adjacent pixels. 
This distribution corresponds to an Airy disk that captures approximately 84\% of the total energy \cite{reagan1993model} and can be mathematically approximated by a two-dimensional Gaussian Point Spread Function (PSF) \cite{lin2012bayesian}.
The radius of the Airy disk exhibits a direct relationship with the standard deviation of the PSF.
First, the standard deviation of the PSF is influenced by the sensor's f-number and the detection wavelength, and it is used to measure energy dispersion. 
Second, the radius of the Airy disk is approximately $1.91$ times the standard deviation of the PSF. 
Furthermore, the Airy disk's radius, known as the Rayleigh Unit (R), defines the sensor’s resolution limit. 
When the inter-target distance falls below $1$R, energy aliasing occurs.
As a result, closely-spaced objects (CSOs) appear as pixel clusters on the imaging plane \cite{macumber2005hierarchical}, making it impossible for the optical system to resolve them as separate points.
The effect is shown in Fig.~\ref{fig:CSIST_PSF}.

In this study, to simulate long-distance infrared small target sequence imaging in a real environment, we implement the 84\% energy concentration definition as the resolution standard, with a diffusion variance of $0.5$ pixel. 
We set the training image size to $11 \times 11$, with \textbf{target counts varying stochastically between $2$, $3$, and $4$ per image}. 
\textbf{Target intensity values are randomly distributed within the range $\textbf{[220, 250]}$}, and the coordinates and intensity values of all targets imaged on the imaging plane (an $11 \times 11$ image) are documented in the corresponding XML files. 
Notably, the target points move along random curves such as quadratic functions, circles, and straight lines.

The SeqCSIST dataset subject to the following constraints:
\begin{enumerate}
    \item \textbf{Spatial Distribution:} All targets are positioned relative to \textbf{a reference point with no intensity value}, with their direction numbers corresponding to targets' numbers (random select in four directions: up, down, left, right). Ensure that \textbf{the direction of the target points relative to the reference point} (from the reference point to the target point) \textbf{remains unchanged during inter-frame motion}.

    \item \textbf{Initial Position and Motion Direction:} The initial coordinate of the reference point in the first frame is denoted as $(x,f(x))$, where $x$ represents an arbitrary integer along the $x$-axis, excluding values that coincide with the image boundaries. \textbf{The direction of the reference point along the trajectory is determined by its initial $x$-coordinate}: if $x<5$, the point progresses along the trajectory in the positive $x$-axis direction; otherwise, it advances in the negative $x$-axis direction.
    
    \item \textbf{Sub-Target Constraints:} The initial distance from each target point to the reference point is set to $0.3$ pixel, meaning the minimum distance between each target point is $0.3\sqrt{2}$ pixel, which ensures that all target points remain within the spatial extent of a single pixel and the targets after unmixing are located at distinct pixels.

    \item \textbf{Inter-Frame Motion Rule:} The reference point moves along the trajectory in consecutive frames, advancing $+0.04$ pixels along the positive x-axis if its initial x-coordinate is less than $5$, or retreating $-0.04$ pixels along the negative x-axis otherwise. Based on this, \textbf{each target point moves a random distance relative to the preceding frame along the direction away from the reference point} by a value in the range $(0, +0.0014)$ pixel.
\end{enumerate}

The overall goal in formulating these rules is to closely replicate the real infrared small target detection process in terms of spatial resolution, target motion patterns, and energy overlap complexity. Below, we explain the design motivation of each rule and its role in achieving a realistic simulation:
\begin{itemize}
    \item \textbf{Initial target distance set to \(0.3\sqrt{2}\) pixels:}  
    This ensures that all targets are confined within a single pixel region. After being processed by the Point Spread Function (PSF) to simulate optical diffraction, this leads to blended energy clusters, faithfully representing the spatial mixing of close small targets in real infrared images.
    \item \textbf{Fixed relative motion direction:}  
    Maintaining a consistent motion direction for targets within a sequence facilitates the simulation of a “target group” moving along a continuous trajectory, which is common in real-world scenarios where multiple targets move coherently.
    \item \textbf{Movement of 0.04 pixels per frame:}  
    This models subpixel-level motion, a critical factor enabling temporal information to aid in energy unmixing. The movement step is chosen small enough to preserve temporal correlation across frames, which reflects the natural continuity of target movement.
    \item \textbf{Four discrete motion directions:}  
    Selecting among four possible directions mimics the random but physically plausible distribution of target movement directions encountered in practical settings, adding realistic variability to the dataset.
\end{itemize}

Following the above principles, the final visual effect of the dataset is shown in Fig.~\ref{fig:dataset}.

The SeqCSIST dataset contains $5,000$ trajectories, comprising a total of $100,000$ frames. Each trajectory consists of $20$ frames, with every $5$ consecutive frames forming one sequence (the initial frame of each sequence has a corresponding sequence number in the trajectory from $0$ to $15$, so there are $16$ sequences in total). The dataset is divided into three subsets: 70\% for training ($3500$ trajectories), 15\% for validation ($750$ trajectories), and 15\% for testing ($750$ trajectories).

%% file: contents/6-method.tex

\section{Method} \label{sec:method}

\subsection{Overview}
\textbf{It should be emphasized that the \textit{unmixing} task proposed in this paper diverges fundamentally from the hyperspectral \textit{unmixing} in remote sensing.} The latter usually involves endmember extraction and abundance estimation under the physical mixing model, while the task proposed in this paper focuses more on \textbf{target separation and sub-pixel positioning.}

The objective of Sequential CSIST Unmixing is to perform sub-pixel localization and unmixing of the targets in the middle frame from an odd number of video frames. 
Consider $2N+1$ video frames $\{L_{t-N}, \ldots, L_{t+N}\}$, where $L_t \in \mathbb{R}^{C \times H \times W}$ denotes the middle frame, while remaining reference frames provide complementary spatiotemporal information for middle frame's targets unmixing. 
The low-resolution frames $\{L_{t-N}, \ldots, L_{t+N}\}$ are processed through the model DeRefNet, yielding a predicted unmixing response $H_t \in \mathbb{R}^{C \times cH \times cW}$, where $c$ is the unmixing ratio.
\begin{equation}
    H_t = f_{DeRefNet}(L_{t-N}, \ldots, L_{t+N}).
    \label{eq:equation_1}
\end{equation}

\begin{figure*}[h]
    \centering
    \includegraphics[width=.98\textwidth]{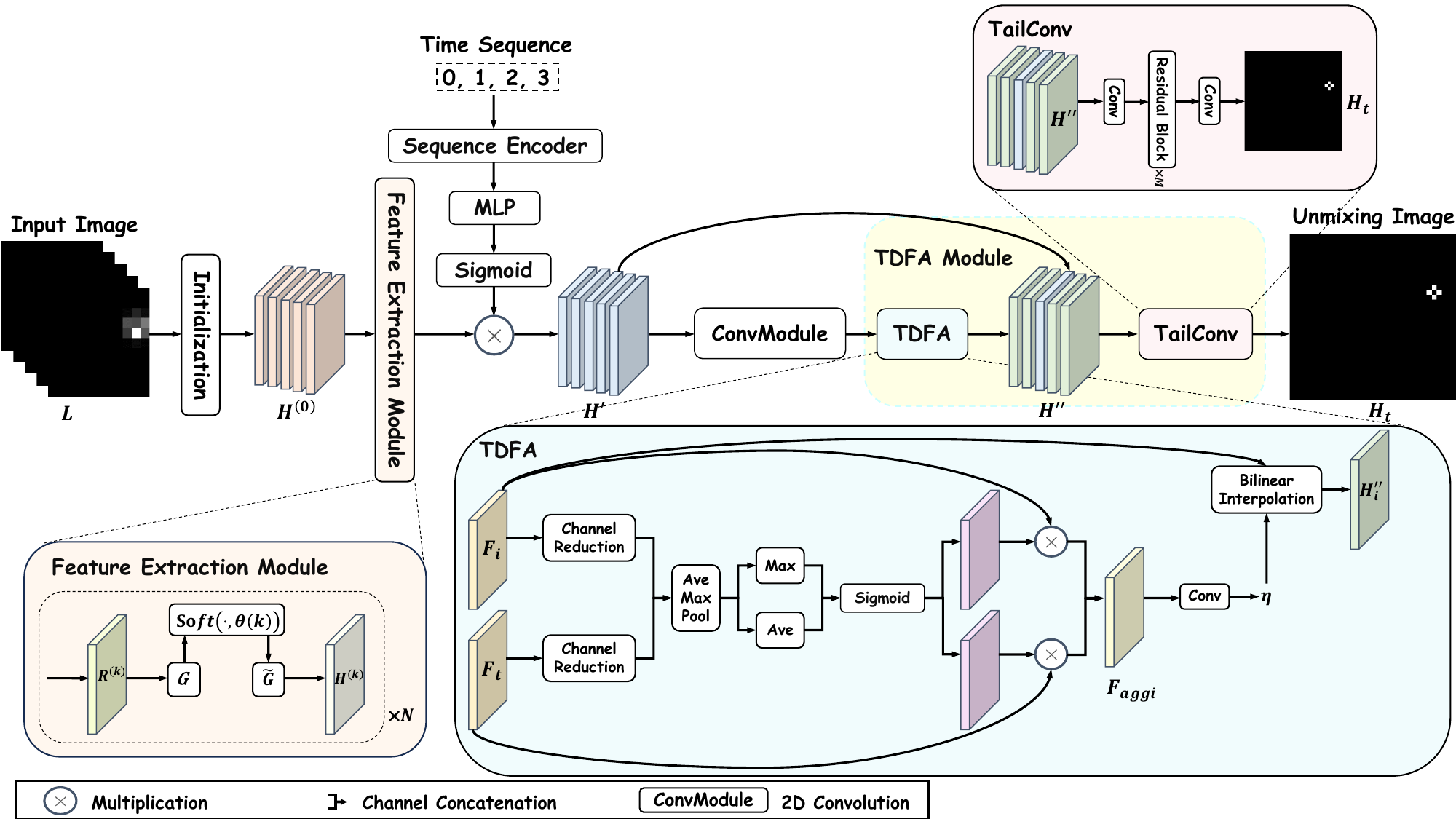}
    \caption{The framework of the proposed DeRefNet approach consists of two main parts. The first part is the feature extraction module, which is used to extract spatial features from the initialized image; subsequently, temporal information is incorporated into the extracted feature map. The second part is the temporal deformable alignment module, which aligns the reference frames to the middle frame.}
    \label{fig:model}
\end{figure*}

The specific network architecture of DeRefNet is shown in Fig.~\ref{fig:model}.
The framework is primarily composed of two main parts: a feature extraction module designed to capture deep-level feature representations from the input sequence, and a Temporal Deformable Feature Alignment (TDFA) module that adaptively aligns the reference frames with the middle frame to ensure precise spatiotemporal correspondence.
Notably, in order to better capture the relationship between frames, we add temporal features to the results after feature extraction.

To emulate realistic optical imaging conditions, GT images of dimension $cm\times cm$ undergo downsampling via sampling matrix $\Phi$ to generate $m\times m$ input frames, where $c$ denotes the downsampling ratio. 
As illustrated in Fig.~\ref{fig:model}, DeRefNet performs target unmixing on the middle frame by processing $m\times m$ sized images from the input sequence. 
Subsequently, these low-resolution frames are initially upscaled to $cm\times cm$ through an initialization matrix $Q_{init}$.
The feature extraction module then captures spatial characteristics across all frames, yielding features that are segregated into $2N$ reference frame features ${H_{t-N}^{(k)}, \ldots, H_{t-1}^{(k)}, H_{t+1}^{(k)}, \ldots, H_{t+N}^{(k)}}$ and middle frame features $H_t^{(k)}$.
The extracted features are augmented with temporal information and then followed by a $2$D convolution module.
Finally, the TDFA module performs spatiotemporal alignment between reference and middle features to guide the target unmixing process, ultimately producing the $H_t$, which represents the unmixed target response.

\subsection{Sampling and Initialization}

To simulate the optical degradation process from high-resolution target distributions to low-resolution measurements, we explicitly construct the downsampling matrix \( \Phi \) based on a continuous-domain point spread function (PSF) model.
Each element in \( \Phi \) quantifies the contribution of a sub-pixel target to a specific low-resolution sensor pixel. Formally, the optical response of a pixel centered at \((x, y)\) to a sub-pixel target located at \((x_t, y_t)\) with brightness \(a_i\) is modeled as a 2D Gaussian diffusion function:
\begin{equation}
    \text{PSF}(x, y; x_t, y_t) = a_i \cdot \frac{1}{2\pi\sigma^2} \exp\left( -\frac{(x - x_t)^2 + (y - y_t)^2}{2\sigma^2} \right).
    \label{eq:equation_1.2}
\end{equation}
To account for sensor integration, the pixel response \(r(x, y)\) is computed by integrating the PSF over the pixel area:
\begin{equation}
    r(x, y) = \iint_{\text{pixel}} \text{PSF}(u, v; x_t, y_t) \, du \, dv.
    \label{eq:equation_1.6}
\end{equation}
This integration is numerically evaluated using the \texttt{dblquad} method in Python, with each low-resolution pixel receiving contributions from a dense grid of simulated sub-pixel targets. The resulting pixel responses from all targets are organized into a matrix \( \Phi \in \mathbb{R}^{m \times n} \), where each row corresponds to a low-resolution sensor element, and each column corresponds to a specific sub-pixel location in the high-resolution grid. The matrix \( \Phi \) effectively captures the spatially-varying blur and integration process that occurs in real optical imaging systems.

According to the above principle, the image obtained on the imaging plane is the low-resolution image, which results from the GT high-resolution image being degraded by the optical system, represented as Eq.~(\ref{eq:equation_1}):
\begin{equation}
    L = \Phi H_{\text{GT}}.
    \label{eq:equation_1}
\end{equation}
The aim of this work is to recover the high-resolution ground truth \( H_{\text{GT}} \) from the low-resolution measurement \( L \), which is similar to solving a Compressed Sensing (CS) problem. Inspired by the traditional method ISTA, DeRefNet first initializes the low-resolution image \( L \). 
Specifically, the initialization method follows~\cite{zhang2018ista}, where a linear mapping \( Q_{\text{init}} \) is used to map the low-resolution \( m \times m \) image \( L_i \) to a high-resolution \( cm \times cm \) image \( H^{(0)} \).

We initialize each high‐resolution estimate \(H^{(0)}\) with a least‐squares mapping for three key reasons: first, by anchoring the network to a physically meaningful starting point, we prevent gradient explosions or vanishing during the early backpropagation steps and ensure training stability; second, this choice preserves the interpretability and convergence guarantees of ISTA by mirroring its analytical initialization, whereas naive convolutional or zero initializations led to slower convergence and reduced robustness; and third, embedding the optimal linear relationship between the low‐resolution input \(L\) and the ground truth \(H_{\text{GT}}\) significantly narrows the effective parameter search space, accelerating network convergence.
The final initialization form is as follows:
\begin{equation}
    Q_{\text{init}} = \arg\min_Q \|QL - H\|^2 = L^T (LL^T)^{-1} H,
    \label{eq:equation_3}
\end{equation}
here \(L = [L_{t-N}, \ldots, L_{t+N}]\) and \(H = [H_{t-N}, \ldots, H_{t+N}]\). 
Geometrically, this projects $L$ onto the subspace spanned by $H_{\text{GT}}$ while minimizing reconstruction bias — a principled alternative to heuristic CNN initialization that empirically amplified high-frequency artifacts.  
This closed-form solution \(L^T(LL^T)^{-1}H\) preserves the algebraic structure of ISTA and ensures that the first network layer starts with the optimal linear unmixing operator.

As shown in Eq.~(\ref{eq:equation_4}), the initialization of \(H_i^{(0)}\) for each input \(L_i\) establishes an affine transformation preserving linear separability in the latent space. This aligns with ISTA’s requirement for convexity in early iterations, ensuring subsequent network layers refine rather than reinvent fundamental mappings.
\begin{equation}
    H_i^{(0)} = Q_{\text{init}} L_i.
    \label{eq:equation_4}
\end{equation}
\subsection{Feature Extraction Module}


To achieve task-specific feature extraction that leverages sparsity priors, we design a spatial feature extraction module that incorporates deep unfolding paragigm within a host network architecture. 
Unlike conventional object detection networks that rely on stacked residual blocks, our approach extracts more semantically meaningful features.
The module transform the initialized $cm \times cm$ feature maps from ${H_{t-N}^{(0)}, \ldots, H_{t+N}^{(0)}}$ to ${H_{t-N}^{(k)}, \ldots, H_{t+N}^{(k)}}$ through Eq. ({\ref{eq:equation_5}}):
\begin{equation}
    H_{t-N}^{(k)}, \ldots, H_{t+N}^{(k)} = f_{FE}(H_{t-N}^{(0)}, \ldots, H_{t+N}^{(0)}).
    \label{eq:equation_5}
\end{equation}

Specifically, the foundation of our approach lies in the canonical ISTA algorithm, which typically solves the CS problem through the following two steps:
\begin{equation}
    R_i^{(k)} = H_i^{(k-1)} - \rho \Phi^T (\Phi H_i^{(k-1)} - L_i),
    \label{eq:equation_6}
\end{equation}
\begin{equation}
    H_i^{(k)} = \arg\min_{H_{\text{GT}i}} \frac{1}{2} \|H_{\text{GT}i} - R_i^{(k)}\|_2^2 + \lambda \|\Psi H_{\text{GT}i}\|_1,
    \label{eq:equation_7}
\end{equation}
Here, $\rho$ is the step size for gradient descent, $\Psi$ is the predefined hand-crafted transformation matrix, and  \(H_i^{(k)}\) represents the k-th iteration of the i-th image, where $i \in \{t-N, \ldots, t+N\}$.

Deep unfolding paradigm extends this framework into a deep architecture with $N$ stages, each corresponding to an ISTA iteration.
Due to the difficulty in solving $H_i^{(k)}$ when dealing with more complex non-orthogonal (and even nonlinear) transformations $\Psi$, and ISTA often requires a large number of iterations to reach the theoretical optimal value, which generally incurs significant computational costs,
deep unfolding paradigm~\cite{zhang2018ista} replaces the hand-crafted matrix $\Psi$ with a trainable nonlinear transformation function ${G}(\cdot)$. 
Given the invertibility ${G}(\cdot)$, its left inverse $\tilde{{G}}(\cdot)$ is defined such that $\tilde{{G}}(\cdot) \circ {G}(\cdot) = \mathbf{I}$. 
The above Eq. ({\ref{eq:equation_7}}) then becomes:
\begin{equation}
H_i^{(k)} = \arg\min_{H_{\text{GT}i}} \frac{1}{2} \|H_{\text{GT}i} - R_i^{(k)}\|_2^2 + \lambda \|{G}(H_{\text{GT}i})\|_1.
\label{eq:equation_8}
\end{equation}
Here, $R_i^{(k)}$ is the direct reconstruction result of $H_i^{(k-1)}$ in the $k$-th iteration, and we expect that in Eq. ({\ref{eq:equation_8}}), the difference between $R_i^{(k)}$ and $H_{\text{GT}i}$ will be minimized. 
Thus, $R^{(k)}$ satisfies the following theorem \cite{zhang2018ista}:
\begin{equation}
    \|{G}(H_{\text{GT}i}) - {G}(R_i^{(k)})\|_2^2 \approx \alpha \|H_{\text{GT}i} - R_i^{(k)}\|_2^2.
    \label{eq:equation_9}
\end{equation}
Through this theorem, Eq. ({\ref{eq:equation_8}}) can be optimized as:
\begin{equation}
    H_i^{(k)} = \arg\min_{H_{\text{GT}i}} \frac{1}{2} \|{G}(H_{\text{GT}i}) - {G}(R_i^{(k)})\|_2^2 + \theta \|{G}(H_{\text{GT}i})\|_1,
    \label{eq:equation_10}
\end{equation}
where $\lambda$ and $\alpha$ are combined into $\theta$, leading to the closed-form version as:
\begin{equation}
    {G}(H_i^{(k)}) = \text{soft}({G}(R_i^{(k)}), \theta)
    \label{eq:equation_11}.
\end{equation}
Given the inverse of ${G}(\cdot)$, $H_i^{(k)}$ can be efficiently computed in closed-form:
\begin{equation}
    H_i^{(k)} = {G}^{-1}(\text{soft}({G}(R_i^{(k)}), \theta)).
    \label{eq:equation_12}
\end{equation}
Here, $\theta$ is a shrinkage threshold and is a learnable parameter. Each stage has its own ${G}(\cdot)$ and $\tilde{G}(\cdot)$ and both of them are learnable processes. Consequently, the above equation can be transformed into:
\begin{equation}
    H_{i}^{(k)} = \tilde{{G}}^{(k)}\left(\text{soft}\left({G}^{(k)}({R}_i^{(k)}), \theta^{(k)}\right)\right).
    \label{eq:equation_13}
\end{equation}
After \(k\)-th iteration, the final feature extraction result \(H_i^{(k)}\) is obtained. Thus, we have Eq. (\ref{eq:equation_5}).

It worth to note that this architecture not only leverages sparsity priors but also enables end-to-end learning of all parameters, resulting in superior feature extraction compared to conventional generic backbone networks, as demonstrated in our ablation studies.

\subsection{Temporal Deformable Feature Alignment Module}
To enhance feature representation across multiple frames, we incorporate temporal information and develop a specialized alignment method that efficiently captures inter-frame relationships. Unlike conventional optical flow methods \cite{dosovitskiy2015flownet} that track complete motion trajectories, our approach selectively aligns key reference frame features with the middle frame through an attention-based architecture, significantly reducing computational complexity while preserving critical information.
\textbf{We first augment the spatial features with temporal information through a learnable encoding process.}
Specially, the fixed position encoding of time information is realized through the position encoding module. 
In order to make better use of time information $t$, we add an MLP layer to realize the learning and utilization of time information by the network, and then use the Sigmoid function to get $T$.
\begin{equation}
    T = Sigmoid(MLP(Encoder(t))).
    \label{eq:equation_14}
\end{equation}
where $t$ represents temporal information and the Sigmoid function ensures proper scaling.
This temporal encoding $T$ is then integrated with the extracted spatial features $H^{(k)}$ via element-wise multiplication:
\begin{equation}
    H^{'} = H^{(k)}T.
    \label{eq:equation_multiple}
\end{equation}
To further enhance representational capacity, we expand the channel dimension of $H^{'}$ through a two-dimensional convolutional layer, generating feature maps $F$, which comprise a middle feature $F_t$ and other reference features ${F_i}$, where $i \in \{t-N, \ldots, t-1, t+1, \ldots, t+N\}$.
\textbf{These features serve as input to the TDFA module}.

\textbf{The TDFA module operates in two complementary stages} to achieve optimal alignment between reference frames and the central frame. Unlike optical flow methods that attempt to track complete motion trajectories explicitly, we achieve adjustable inter-frame alignment by selective attention, \textbf{an implicitly learnable strategy which leverages sparsity priors to focus only on high-activation regions—those with the strongest features}.
In the first stage, we implement a selective attention architecture that dynamically combines reference and middle frame features:
First, the reference frame and the middle frame features are dynamically combined by selective attention process. 
\begin{equation}
    F_{aggi} = SelectiveAttention(F_{i}, F_{t}).
    \label{eq:equation_multiple}
\end{equation}
This process begins by reducing the channel dimensions of both feature maps through convolution, followed by their concatenation. We then extract maximum and average pooled features, activated through a Sigmoid function. Recognizing the central frame's primacy in our processing pipeline, we strategically weight the attention: average-pooled weights are applied to the central frame features, while maximum-pooled weights are applied to reference frames, emphasizing their most salient characteristics. The resulting weighted features are concatenated along the channel dimension to create the aggregated feature map $F_{aggi}$.

The second stage adaptively extracts features from the reference frame by assessing their correlation with the aggregated features $F_{aggi}$, which enables the model to effectively align and integrate complementary information from the reference frame with the overall feature representation.
The process predicts sampling parameters $\eta$ for each reference feature map $F_{i}$:
\begin{equation}
    \eta = conv(F_{aggi}).
    \label{eq:equation_15}
\end{equation}
Here, $\eta = \{\Delta p_k\}$ contains content-dependent offsets matrices employed to adjust the sampling window to better capture complex structures of aggregated features $F_{aggi}$.
Using these learned offsets, we modulate each reference feature map $F_{i}$ into an aligned representation $H_{i}^{''}$ through a bilinear interpolation $\phi{(\cdot)}$.
\begin{equation}
    H_{i}^{''} = \phi(F_{i}, \eta)
    \label{eq:equation_16}
\end{equation}
For any spatial location $P$ in $H_{i}^{''}$, the feature response is computed as:
\begin{equation}
    H_{i}^{''}(p) = \sum_{k=1}^{K^2} \omega_k \cdot F_{i}(p + p_k + \Delta p_k).
    \label{eq:equation_17}
\end{equation}
Here, $K^2$ is the total number of sampling locations in the grid, $\omega_k$ represents the weight assigned to the $k$-th sampling location, $p_k$ denotes the fixed sampling offset, and $\Delta p_k$ is the dynamic offset that adjusts the fixed sampling location based on the input features, which is $\eta = \{\Delta p_k\}$. Note that $\Delta p_k$ can be a fractional value and to manage this, bilinear interpolation is used, following the approach described in \cite{dai2017deformable}, to accurately compute the feature values at these non-integer locations.

Finally, the aligned $2N$ reference features $H_{i}^{''}$ and one middle frame feature $H_t^{(k)}$ are aggregated through tail convolution to obtain the unmixed image $H_t$. To enhance the network's representational capacity and optimization stability, this aggregation process contains a cascade of $n$ residual blocks, which effectively deepens the architecture while mitigating gradient vanishing through skip connections.

\subsection{Loss Function}
The loss of our DeRefNet consists of three parts: constraint loss, alignment loss, and regression loss. Specifically, the constraint loss ensures the reversibility of the internal process \( \tilde{G}(\cdot) \circ G(\cdot) = \mathbf{I} \) during each iteration within the feature extraction module. Its purpose is to reduce the difference between the initialized image and the extracted feature image, which is specifically formulated as shown in Eq. (\ref{eq:equation_18}):
\begin{equation}
    \mathcal{L}_{\text{constraint}} = \frac{1}{X Y} \sum_{i=1}^{X} \sum_{i=1}^{L} \left\| \tilde{{G}}^{(k)}\left({G}^{(k)}(s_i)\right) - s_i \right\|_2^2.
    \label{eq:equation_18}
\end{equation}
Here $X$ is the size of the training set, $L$ is the number of stages, $Y$ is the size of the super-resolved image feature, and $s_i$ represents the GT. 

The alignment loss refers to the loss generated from aligning the reference frame with the middle frame. Reducing the alignment loss ensures that the difference between the aligned reference frame and the middle frame within the temporal deformable feature alignment block is minimized. The specific formulation of the alignment loss is as follows:
\begin{equation}
    \mathcal{L}_{\text{align}} = \frac{1}{2N} \sum_{i=t-N, i\neq t}^{t+N} \left\| H_{i}^{''} - H_{t}^{'} \right\|_1.
    \label{eq:equation_19}
\end{equation}
The regression loss represents the detection difference between the model's output and the GT. The specific formulation is as follows:
\begin{equation}
    \mathcal{L}_{\text{regression}} = \frac{1}{(T-4)*(M/T)} \sum_{k=1}^{M/T} \sum_{i=2}^{T-2} \left\| H_{ki} - s_{ki} \right\|_2^2,
    \label{eq:equation_20}
\end{equation}
Here, $T$ represents the frame numbers of a trajectory, $H_{ki}$ and $s_{ki}$ represent the reconstructed result of the $i$-th frame of the $k$-th trajectory and its corresponding GT, respectively, $M/T$ represents the number of trajectories.

Combining the above three losses, the total loss of the DeRefNet model is obtained as follows:
\begin{equation}
    \mathcal{L}_{all} = \beta \mathcal{L}_{constraint} + \gamma \mathcal{L}_{align} + \zeta \mathcal{L}_{regression}
    \label{eq:equation_21}
\end{equation}
where $\beta$, $\gamma$ and $\zeta$ are the weight coefficients for each loss term, and we set $\beta$ and $\gamma$ to 0.01, $\zeta$ to 1.
To validate the choice of these hyperparameters, we conduct a grid search experiment with six different weight configurations:
\begin{itemize}
\item \textbf{Group A (Default)}: $\zeta=1$, $\beta=0.01$, $\gamma=0.01$ — Balanced supervision; used as baseline.
\item \textbf{Group B}: $\zeta=1$, $\beta=0.05$, $\gamma=0.05$ — Weaken auxiliary constraints; assess dominance of regression loss.
\item \textbf{Group C}: $\zeta=1$, $\beta=0.2$, $\gamma=0.2$ — Amplify auxiliary signals; test whether stronger regularization improves generalization.
\item \textbf{Group D}: $\zeta=1$, $\beta=0.2$, $\gamma=0.05$ — Emphasize transform constraint while keeping alignment weak.
\item \textbf{Group E}: $\zeta=1$, $\beta=0.05$, $\gamma=0.2$ — Emphasize temporal alignment; test its contribution to overall accuracy.
\item \textbf{Group F}: $\zeta=0.5$, $\beta=0.1$, $\gamma=0.1$ — Downweight main loss to observe impact of auxiliary dominance.
\end{itemize}
\begin{table}[htbp]
  \renewcommand\arraystretch{1.2}
  \footnotesize
  \centering
  \caption{Grid Experiment on Loss Weight Configurations: \\ Balancing Main Regression Loss with Auxiliary Constraints for Optimal CSIST Detection}
  \setlength{\tabcolsep}{4pt}
  \begin{tabular}{ccccccc}
  \toprule
  \textbf{Group} & \textbf{CSO-mAP} & \textbf{AP$_{05}$} & \textbf{AP$_{10}$} & \textbf{AP$_{15}$} & \textbf{AP$_{20}$} & \textbf{AP$_{25}$} \\
  \midrule
  A & \textbf{51.55} & \textbf{1.00} & \textbf{14.40} & \textbf{54.90} & 90.40 & 97.10 \\
  B & 50.82 & 0.70 & 11.00 & 51.50 & 92.20 & 98.60 \\
  C & 50.44 & 0.70 & 10.40 & 50.50 & 92.00 & 98.60 \\
  D & 50.54 & 0.70 & 10.50 & 50.70 & 92.10 & 98.80 \\
  E & 51.11 & 0.80 & 11.40 & 51.90 & 92.50 & 98.90 \\
  F & 50.86 & 0.80 & 10.90 & 51.10 & \textbf{92.60} & \textbf{99.00} \\
 \bottomrule
 \label{table:Grid Experiment}
\end{tabular}
\vspace{-1\baselineskip}
\end{table}
The experimental results demonstrate that our default configuration (Group A) achieves the best overall performance. The superior performance of Group A validates our hyperparameter selection rationale: (1) Setting $\zeta=1$ ensures the regression loss maintains its primary role in learning temporal boundaries; (2) The modest weights $\beta=\gamma=0.01$ for auxiliary losses provide sufficient regularization without overwhelming the main objective. Groups B-F show that either weakening auxiliary constraints (Group B) or over-emphasizing them (Groups C-D) leads to degraded performance in critical metrics, while reducing the main loss weight (Group F) compromises precise localization despite marginal improvements in loose temporal thresholds. This ablation confirms that our balanced approach optimally leverages both primary supervision and auxiliary regularization.

%% file: contents/7-experiment.tex

\section{Experiments} \label{sec:experiment}

\subsection{Experimental Settings} \label{subsec:setting}
\textbf{Metrics:} To evaluate model performance in the Sequential CSIST Unmixing task, we adopt the \textbf{CSO mean Average Precision (CSO-mAP)}, a customized assessment metric for this task.


The evaluation process begins by categorizing each prediction into either True Positives (TP) or False Positives (FP), and a binary list is created in line with the conventions of COCO. 
In this list, TP predictions are represented by 1, while FP predictions are denoted by 0. This binary representation forms the basis for constructing the Precision-Recall (PR) curve. 
By systematically adjusting the confidence threshold for positive predictions, multiple pairs of Precision and Recall values are derived, which collectively trace out the PR curve. The Average Precision (AP), computed as the area under the PR curve, offers a detailed assessment of the model's behavior across varying confidence levels, capturing the balance between Precision and Recall. To further standardize evaluation, we use the CSO mean Average Precision (CSO-mAP), which aggregates AP scores over different distance thresholds \(\delta_k \). This metric provides a consistent framework for comparing model performance in the context of the Sequential CSIST Unmixing task.

\textbf{Training Settings:} As a downstream post-processing task of object detection, our method processes target patches cropped from detection results rather than raw input images. 
Throughout the experiment, the initialized super-resolution ratio $c$ is set to $3$. The input target patches size is $11 \times 11$, and both the GT image size and output unmixing image size are $33 \times 33$. 
In order to load a complete trajectory at each time, the batch size is set to $20$ in our experiments.
To improve data utilization and achieve better model performance, $5$ consecutive frames are processed at a time ($1-5, 2-6, ..., 16-20$). Therefore, each trajectory contains $20 - 4 = 16$ sequences, instead of $20 \div 5 = 4$ sequences. 
Finally, the number of residual blocks $n$ is set to $5$.
In all parts of the experiment involving spatial feature extraction, a $32$-channel convolution is used. 
For optimization, we employ the Adam optimizer and the MMEngine framework to execute the network, with the learning rate consistently maintained at $10^{-4}$.

\subsection{Ablation Study} \label{subsec:ablation}

\textbf{1) Impact on Early-stage Feature Extraction:} 
To evaluate the effectiveness of deep unfolding paradigm against conventional stacked residual blocks, we conducted a controlled ablation study using the multi-frame feature extraction paradigm \cite{tian2020tdan}. By substituting the residual stacking structure at the network's head with our deep unfolding architecture, we isolated the impact of this architectural choice on feature extraction performance.

As shown in Table \ref{Performance of Feature Extraction Module and stacked residual blocks structure}, the deep unfolding approach consistently outperformed the baseline across all evaluation metrics. The performance gains were particularly evident across multiple IoU thresholds, demonstrating the robustness of our approach under varying precision requirements.

This consistent improvement can be attributed to two key factors. First, unlike generic residual stacking, our deep unfolding architecture incorporates task-specific sparsity priors, enabling more targeted feature extraction that aligns with the inherent characteristics of infrared imagery. Second, by shifting the super-resolution process to the front of the pipline, our approach effectively addresses the blending issues that commonly affect small targets in infrared scenes.

The particularly significant improvements observed at lower IoU thresholds highlight the deep unfolding architecture's enhanced ability to discriminate between closely spaced targets—a critical capability in densely populated scenarios where detection conflicts frequently occur. These results provide compelling evidence that task-specific architectures incorporating appropriate prior knowledge outperform generic feature extraction methods, even when the latter have been extensively optimized for similar tasks.

\begin{table}[H]
  \renewcommand\arraystretch{1.2}
  \footnotesize
  \centering
  \caption{Performance of Feature Extraction Module and stacked residual blocks structure}
  \setlength{\tabcolsep}{4pt}
  \begin{tabular}{ccccccc}
  \toprule
  \textbf{Backbone} & \textbf{CSO-mAP} & \textbf{AP$_{05}$} & \textbf{AP$_{10}$} & \textbf{AP$_{15}$} & \textbf{AP$_{20}$} & \textbf{AP$_{25}$} \\
  \midrule
  ResBlocks \cite{tian2020tdan}   & 47.96 & 0.50 & 8.60 & 43.80 & 89.30 & 97.50  \\
  Deep Unfolding & \textbf{\underline{50.27}} & \textbf{\underline{0.70}} & \textbf{\underline{11.40}} & \textbf{\underline{50.80}} & \textbf{\underline{90.60}} & \textbf{\underline{97.90}} \\
 \bottomrule
 \label{Performance of Feature Extraction Module and stacked residual blocks structure}
\end{tabular}
\vspace{-1\baselineskip}
\end{table}

\textbf{2) Impact on Deformable Alignment:} To systematically evaluate the superiority of the Deformable Alignment (DA) module over the traditional optical flow method, we conducted a comparative performance evaluation of both modules. 
The performance of both modules was quantitatively compared across multiple thresholds, as shown in Table \ref{table:Performance of different Modules}.

The results of the ablation experiment reveal that the deformable alignment module, utilizing interpolation-based deformable sampling, significantly outperforms the traditional optical flow method in taget detection across various thresholds. 
It consistently achieves higher average precision (AP) across multiple IoU thresholds compared to optical flow methods, revealing the latter's inherent limitations in capturing displacement information within CSIST scenarios. Notably, the deformable module maintains robust AP performance even at lower IoU thresholds, attributable to its dynamic receptive field adjustment mechanism that effectively adapts to complex motion patterns and spatial variations.
This shows that the module can still maintain its superiority when the target distance is very close.

\textbf{3) Impact on Time Encoder:} 
As shown in the second and third rows of Table~\ref{table:Performance of different Modules}, to highlight the importance of timing information, we add a time encoder based on the multi-frame unmixing framework.

The results indicate that adding the time encoder enhances model performance across multiple IoU thresholds. Specifically, while both models use deformable alignment to improve spatial feature extraction, the inclusion of the time encoder further refines motion representation, leading to more robust and temporally aware predictions. Notably, AP scores improve consistently at higher IoU thresholds, demonstrating that temporal encoding helps the model capture sequential dependencies and motion patterns more effectively.
This suggests that the time encoder compensates for the limitations of static spatial alignment by introducing richer temporal context, ultimately leading to improved performance in dynamic environments.

\begin{table*}[htbp]
  \renewcommand\arraystretch{1.2}
  \footnotesize
  \centering
    \vspace{-1\baselineskip}
  \caption{Systematic Module-wise Ablation Study: \\ Impact Assessment of Core Components on DeRefNet Performance and Computational Cost}
  \begin{tabular}{ccccc|cc|cccccc}
  \toprule
  \textbf{Deep Unfolding} & \textbf{Optical Flow} & \textbf{DA} & \textbf{Time Encoder} & \textbf{DDA} & \textbf{Paras} & \textbf{FLOPs} & \textbf{CSO-mAP} & \textbf{AP$_{05}$} & \textbf{AP$_{10}$} & \textbf{AP$_{15}$} & \textbf{AP$_{20}$} & \textbf{AP$_{25}$} \\
  \midrule
  \checkmark & \checkmark & - & - & - & \makecell{/} & \makecell{/} & 50.55 & 0.70 & 11.20 & 49.50 & 92.50 & 98.80  \\
  \checkmark & - & \checkmark & - & - & 0.23 M & 6.44 G & 50.67 & 0.80 & 11.90 & 50.90 & 91.50 & 98.30 \\
  \checkmark & - & \checkmark & \checkmark & - & \makecell{/} & \makecell{/} & 51.39 & 0.80 & 12.40 & 52.20 & 92.70 & 98.80 \\
  \checkmark & - & - & - & \checkmark & 0.28 M & 6.14 G & 51.09 & 0.80 & 12.00 & 52.00 & 92.30 & 98.40 \\
  \bottomrule
\end{tabular}
\label{table:Performance of different Modules}
\vspace{-1\baselineskip}
\end{table*}

\textbf{4) Impact on Dynamic Deformable Alignment:} 
To improve performance and reduce model weight, we replaced the deformable stacking structure with a selection attention structure, creating the Dynamic Deformable Alignment (DDA) module.
As shown in the second and fourth rows of Table~\ref{table:Performance of different Modules}, the experimental results support this intent: while Experiment 2 (using the deformable stacking structure) registers 6.44G FLOPs, Experiment 4 (using the selection attention mechanism) achieves superior performance with only 6.14G FLOPs. 
This reduction in computational cost demonstrates that the selection attention method improves model performance while streamlining the architecture for enhanced efficiency. By dynamically prioritizing salient features, it bolsters feature representation and alignment precision, leading to more robust detection outcomes. These benefits highlight the proposed module as a lightweight yet powerful alternative to traditional deformable stacking structures.

\begin{table}[H]
  \renewcommand\arraystretch{1.2}
  \footnotesize
  \centering
  \caption{DeRefNet performance on SeqCSIST and Hybrid datasets with Hybrid referring to simulated CSOs composited onto real-world infrared scenes}
  \setlength{\tabcolsep}{4pt}
  \begin{tabular}{ccccccc}
  \toprule
  \textbf{DATASET} & \textbf{CSO-mAP} & \textbf{AP$_{05}$} & \textbf{AP$_{10}$} & \textbf{AP$_{15}$} & \textbf{AP$_{20}$} & \textbf{AP$_{25}$} \\
  \midrule
  SeqCSIST & \textbf{51.55} & \textbf{1.00} & \textbf{14.40} & \textbf{54.90} & \textbf{90.40} & \textbf{97.10} \\
  Hybrid Dataset & 47.48 & 0.70 & 11.00 & 46.30 & 84.80 & 94.60 \\
 \bottomrule
 \label{tab:cross_domain_results}
\end{tabular}
\vspace{-1\baselineskip}
\end{table}

\textbf{5) Cross-Domain Generalization Validation:} 
To evaluate the robustness and generalization capability of our proposed DeRefNet under domain shift conditions, we conducted a supplemental experiment using a hybrid dataset that simulates realistic deployment scenarios.

Specifically, we superimposed the 5,000 target trajectories from the synthetic SeqCSIST dataset onto 5,000 real-world infrared background sequences. These real backgrounds were collected from diverse scenes including mountains, forests, buildings, towers, and atmospheric environments, thus exhibiting more complex noise patterns and background textures than those present in the synthetic domain.
This hybrid dataset introduces realistic variations while maintaining ground-truth unmixing annotations, allowing for a meaningful cross-domain evaluation.

We trained and tested DeRefNet on this hybrid dataset using the same training protocol as on SeqCSIST. As the comparative results shown in Table~\ref{tab:cross_domain_results}, the mAP of DeRefNet drops by only 4.07\% when applied to the hybrid dataset, indicating that the model retains strong detection and unmixing capability even under complex real-scene backgrounds. This result highlights the effectiveness and robustness of DeRefNet in handling realistic domain variations.

While the absence of publicly available real annotated unmixing datasets still limits full real-world evaluation, this experiment provides a meaningful intermediate validation step. In future work, we plan to explore domain adaptation strategies to further enhance cross-dataset transferability.

\begin{table}[htbp]
  \renewcommand\arraystretch{1.2}
  \footnotesize
  \centering
  \caption{Comprehensive Noise Tolerance Assessment: DeRefNet Performance Analysis Under Additive Gaussian Noise with Different Standard Deviation Values}
  \setlength{\tabcolsep}{4pt}
  \begin{tabular}{ccccccc}
    \toprule
    \(\sigma\) & CSO-mAP & AP$_{05}$ & AP$_{10}$ & AP$_{15}$ & AP$_{20}$ & AP$_{25}$ \\
    \midrule
    0 (clean) & \textbf{51.55} & \textbf{1.00} & \textbf{14.40} & \textbf{54.90} & \textbf{90.40} & \textbf{97.10} \\
    2         & 49.09         & 0.80          & 12.40         & 50.50         & 86.60          & 95.30          \\
    3         & 48.40         & 0.80          & 11.80         & 48.00         & 86.00          & 95.30          \\
    4         & 48.41         & 0.90          & 12.20         & 48.00         & 85.80          & 95.20          \\
    5         & 47.23         & 0.80          & 11.40         & 46.10         & 84.10          & 93.90          \\
    \bottomrule
    \label{table:noise_robustness}
  \end{tabular}
\end{table}

\textbf{6) Robustness to Additive Gaussian Noise:} To assess DeRefNet’s resilience against sensor and thermal noise, we synthetically perturbed the input frames with zero-mean Gaussian noise of varying intensity (\(\sigma=2\)–5) while keeping the network architecture unchanged.  This noise range corresponds to signal-to-noise ratios (33–42 dB) typical of infrared imaging systems.  We exclude \(\sigma<1\) (too weak to challenge the model) and \(\sigma>10\) (unrealistic distortion), focusing on \(\sigma=2\)–5 as representative mid-level noise.

As shown in Table~\ref{table:noise_robustness}, even under moderate noise (\(\sigma=2\)), DeRefNet’s mAP drops by only 2.46\%, and with stronger perturbation (\(\sigma=5\)), it still achieves mAP \(\approx\) 47.2\% and AP$_{25}$ \(\approx\) 93.9\%. 
These results demonstrate that our DeRefNet model can effectively resist signal-level interference and maintain good generalization performance under realistic noisy conditions.

\begin{table}[H]
  \renewcommand\arraystretch{1.2}
  \footnotesize
  \centering
  \caption{Multi-Target Unmixing Performance Evaluation: DeRefNet Effectiveness Analysis Under Variable Target Population Densities in Sequential Data Processing}
  \setlength{\tabcolsep}{4pt}
  \begin{tabular}{ccccccc}
  \toprule
  \textbf{Dataset} & \textbf{CSO-mAP} & \textbf{AP$_{05}$} & \textbf{AP$_{10}$} & \textbf{AP$_{15}$} & \textbf{AP$_{20}$} & \textbf{AP$_{25}$} \\
  \midrule
  SeqCSIST (2-4) & \textbf{51.55} & \textbf{1.00} & \textbf{14.40} & \textbf{54.90} & \textbf{90.40} & \textbf{97.10} \\
  SeqCSIST (2-8) & 49.79 & 0.70 & 12.60 & 50.60 & 87.90 & 97.10 \\
 \bottomrule
 \label{table:DeRefNet_2_8}
\end{tabular}
\vspace{-1\baselineskip}
\end{table}

\begin{table*}[htbp]
  \renewcommand\arraystretch{1.4}
  \footnotesize
  \centering
  \caption{Systematic Benchmarking Analysis of DeRefNet Performance: Comparative Evaluation Against Multiple Method Categories Including Classical Optimization, Super-Resolution Networks, and Deep Unfolding Architectures on \textbf{SeqCSIST} Dataset.}
  \label{table:Comparison with State-of-the-Arts methods on SeqCSIST dataset}
  \vspace{-2pt}
  \setlength{\tabcolsep}{6.pt}
  \begin{tabular}{l|ccc|cccccc}
  \multirow{2}{*}{\textbf{Method}} & \multirow{2}{*}{\textbf{FPS} $\uparrow$}   & \multirow{2}{*}{\textbf{Params} $\downarrow$} & \multirow{2}{*}{\textbf{FLOPs} $\downarrow$} & \multicolumn{6}{c}{CSO-mAP $\uparrow$}   \\
  & & & & {\textbf{CSO-mAP}} & \textbf{AP$_{05}$} & \textbf{AP$_{10}$} & \textbf{AP$_{15}$} & \textbf{AP$_{20}$} & \textbf{AP$_{25}$} \\
  \Xhline{1pt}
  \multicolumn{9}{l}{\fontsize{10.5}{13}\selectfont \textit{Traditional Optimization}} \\ \hline
  ISTA~\cite{Daubechies2004ISTA} & 0.1 &  & 398.57 M & 10.72  & 0.14  & 1.97  & 8.74  & 18.22 & 24.53 \\
  \hline
  BID~\cite{levin2007blind} & 0.1 & & 10.89 M & 14.40  & 0.00  & 3.00  & 13.00  & 26.00 & 30.00 \\
  \hline
  \multicolumn{9}{l}{\fontsize{10.5}{13}\selectfont \textit{Image Super-Resolution}}  \\
  \hline
  SRCNN~\cite{dong2015image} & 102{,}961 & 15.84 K & 0.35 G & 49.64 & 1.40 & 16.30 & 51.20 & 85.00 & 94.30 \\
  \hline
  GMFN~\cite{li2019gated} & 855 & 2.80 M & 27.53 G & 50.94 & 0.70 & 11.90 & 51.20 & 92.10 & 98.80 \\
  \hline
  DBPN~\cite{haris2018deep} & 7{,}109 & 1.96 M & 4.75 G & 50.40 & 0.80 & 12.50 & 51.20 & 90.00 & 97.40 \\
  \hline
  SRGAN~\cite{ledig2017photo} & 12{,}965 & 35.31 M & 40.27 G & 26.96 & 0.30 & 3.90 & 19.40 & 46.90 & 64.30 \\
  \hline
  BSRGAN~\cite{zhang2021designing} & 1{,}528 & 36.06 M & 0.27 T & 33.21 & 0.40 & 6.10 & 27.50 & 57.20 & 74.90 \\
  \hline
  ESRGAN~\cite{wang2018esrgan} & 1{,}024 & 50.45 M & 0.38 T & 36.86 & 0.40 & 6.00 & 30.30 & 66.80 & 80.70 \\
  \hline
  RDN~\cite{zhang2018residual} & 919 & 22.31 M & 53.97 G & 49.61 & 0.70 & 10.60 & 48.20 & 90.40 & 98.20 \\
  \hline
  EDSR~\cite{lim2017enhanced} & 11{,}476 & 0.39 M & 0.99 G & 50.19 & 0.60 & 10.30 & 48.80 & 92.20 & 99.00 \\
  \hline
  ESPCN~\cite{shi2016real} & 144{,}901 & 54.75 M & 22.73 K & 47.18 & 1.60 & 15.30 & 46.60 & 80.30 & 92.00 \\
  \hline
  TDAN~\cite{tian2020tdan} & 259 & 0.59 M & 2.18 G & 47.96 & 0.50 & 8.60 & 43.80 & 89.30 & 97.50 \\
  \hline
  \multicolumn{9}{l}{\fontsize{10.5}{13}\selectfont \textit{Deep Unfolding}}  \\
  \hline
  LIHT~\cite{wang2016learning} & 253 & 21.10 M & 0.42 G & 6.36 & 0.10 & 1.00 & 4.30 & 10.40 & 16.00 \\
  \hline
  LAMP~\cite{metzler2017learned} & 7{,}172 & 2.13 M & 86.97 G & 9.09 & 0.10 & 1.50 & 6.50 & 15.00 & 22.30 \\
  \hline
  ISTA-Net~\cite{zhang2018ista} & 4{,}052 & 0.17 M & 4.09 G & 48.95 & 0.70 & 11.20 & 49.70 & 87.70 & 95.40 \\
  \hline
  FISTA-Net~\cite{xiang2021fista} & 4{,}052 & 74.60 K & 6.02 G & 50.61 & 1.00 & 12.60 & 51.40 & 90.70 & 97.30 \\
  \hline
  ISTA-Net+~\cite{zhang2018ista} & 5{,}504 & 0.38 M & 7.70 G & 51.02 & 1.00 & 13.70 & 52.70 & 90.40 & 93.70 \\
  \hline
  ISTA-Net++~\cite{you2021ista} & 1{,}751 & 0.76 M & 16.54 G & 50.50 & 0.70 & 10.40 & 49.20 & 92.8 & 99.40 \\
  \hline
  LISTA~\cite{gregor2010learning} & 490 & 21.10 M & 0.42 G & 9.39 & 0.10 & 1.70 & 6.90 & 15.40 & 22.70 \\
  \hline
  USRNet~\cite{zhang2020deep} & 622 & 1.07 M & 11.26 G & 49.25 & 0.70 & 9.80 & 46.60 & 91.20 & 98.90 \\
  \hline
  TiLISTA~\cite{liu2019alista} & 4{,}716 & 2.22 M & 86.97 M & 13.52 & 0.20 & 2.10 & 9.50 & 22.60 & 33.30 \\
  \hline
  RPCANet~\cite{wu2024rpcanet} & 2{,}601 & 0.68 M & 14.81 G & 47.17 & 0.70 & 10.20 & 44.50 & 84.60 & 95.90 \\
  \hline
  \rowcolor[rgb]{0.9,0.9,0.9} $\star$ \textbf{DeRefNet (Ours)} & 367 & 0.89 M & 15.70 G & \textbf{\underline{51.55}} & 1.00 & 14.40 & 54.90 & 90.40 & 97.10 \\
  \end{tabular}
  \vspace{-1\baselineskip}
\end{table*}

\textbf{7) Adaptability to Dynamic Target Quantities:} To assess DeRefNet's robustness in scenarios exceeding its training distribution of 2–4 co-occurring targets per patch, we conducted targeted experiments to evaluate DeRefNet's performance under increased target densities. We extended the target range from 2-4 to 2-8 targets per sequence, which represents a realistic upper bound for close-range infrared scenarios based on domain expertise. Sequences with 5-8 targets introduce significant challenges including increased inter-target energy aliasing, spatial interference, and computational complexity.
The experimental protocol is as follows:
\begin{itemize}
    \item Generated extended SeqCSIST dataset with 2-8 targets per sequence
    \item Maintained identical training protocols and hyperparameters
    \item Evaluated using the same metrics for direct comparison
\end{itemize}
As shown in the Table~\ref{table:DeRefNet_2_8}, the results reveal several important findings regarding DeRefNet's adaptability to increased target densities: 1) DeRefNet experiences a 1.76\% decrease in CSO-mAP (from 51.55\% to 49.79\%) when target density increases from 2-4 to 2-8 range. While this represents a moderate performance decline, it demonstrates that the model maintains reasonable detection capability under more challenging conditions. 2) In the case of low and medium IoU Thresholds, although the unmixing accuracy is reduced, the consistency of the indicators can still be maintained. Moderate performance drop at {AP$_{20}$} (-2.5\%), while {AP$_{25}$} remains stable (97.10\%), indicating that when the model achieves high-confidence detections, precision is preserved.

These findings substantiate two critical advantages of our temporal unmixing framework: 1) Effective generalization beyond training distribution boundaries through robust feature representation learning, and 2) Maintenance of both unmixing fidelity and spatial localization precision under dynamic target population conditions. The results particularly highlight the architecture's suitability for real-world applications where target density variations are inherently unpredictable.

\begin{figure*}[b]
    \centering
    \includegraphics[width=.98\textwidth]{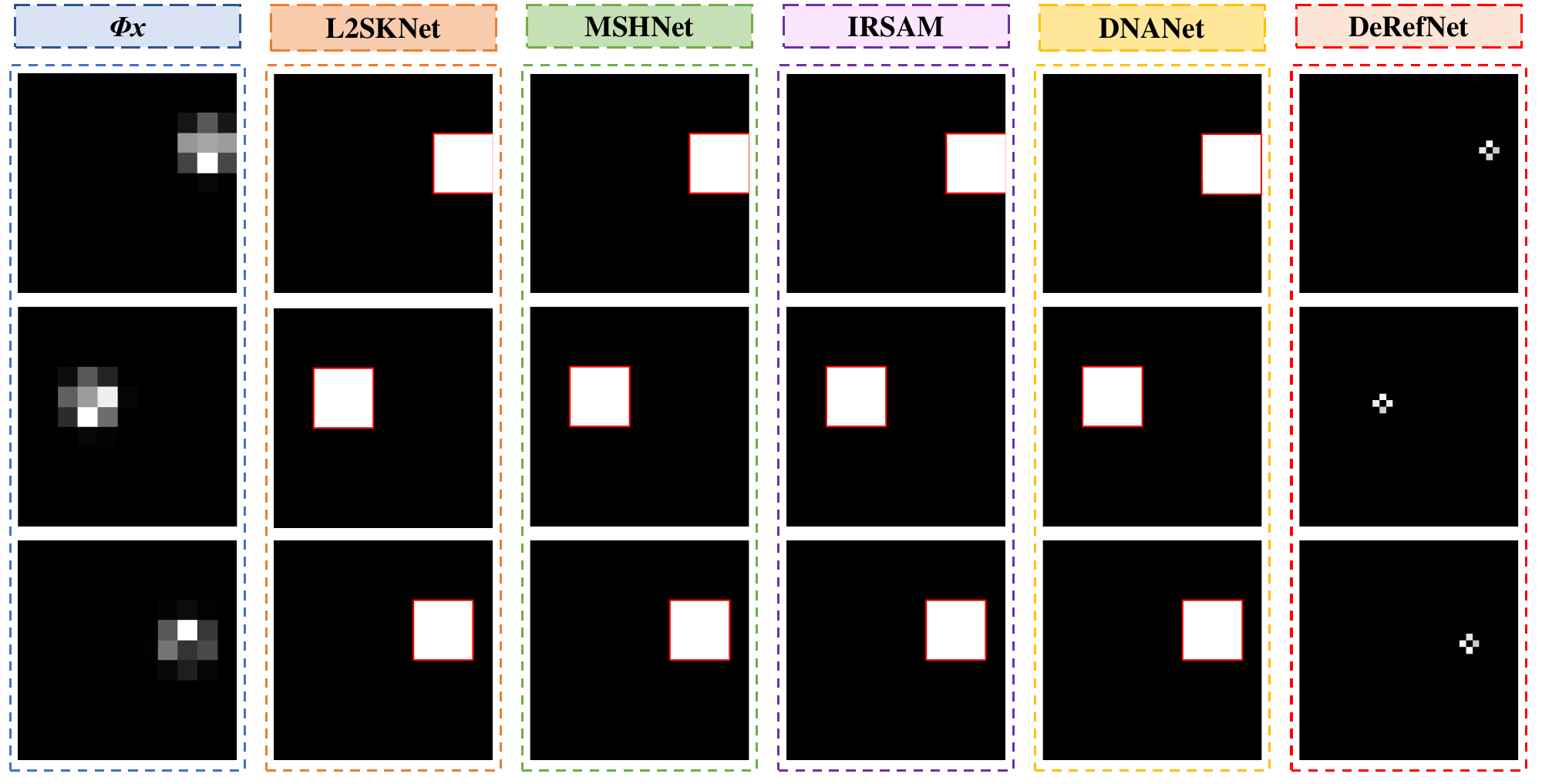}
    \caption{Partial frame unmixing effect of subsequences of different trajectories. The visualization demonstrates DeRefNet's superior capability in resolving closely-spaced infrared small targets, where individual target signatures and precise sub-pixel positions are successfully extracted from densely clustered regions that conventional detection methods can only provide as coarse binary mask results.}
    \label{fig:comparison}
\end{figure*}

\subsection{Comparison with State-of-the-Arts} \label{subsec:sota}

To validate the effectiveness of the model, we compared our DeRefNet on dataset SeqCSIST with other methods, including ISTA \cite{gregor2010learning} and ISTA-Net \cite{zhang2018ista}, among others. The specific experimental results are shown in Table \ref{table:Comparison with State-of-the-Arts methods on SeqCSIST dataset}.

From the data in Table \ref{table:Comparison with State-of-the-Arts methods on SeqCSIST dataset}, we can conduct a detailed analysis and comparison of the performance of different models at various IoU thresholds for AP. 
The experimental results demonstrate that \textbf{DeRefNet} exhibits exceptional target unmixing capabilities on the SeqCSIST dataset, outperforming other methods in terms of mAP. 
We conducted the experiment in \textbf{three} parts, and the specific results are as follows:

\textbf{1) Compared to traditional model-driven optimization methods} (\textit{e.g.}, ISTA and BID), DeRefNet significantly outperforms them in the CSIST scenario, achieving an mAP of 51.55, much higher than ISTA (10.72) and BID (14.40). This substantial disparity highlights the limitations of purely model-driven approaches in effectively handling Sequential CSIST Unmixing. In contrast, DeRefNet, by combining both model-driven and data-driven strategies, excels in resolving overlapping targets and adapting to diverse scene conditions, offering a significant advantage in unmixing and localization accuracy.

\textbf{2) Compared to super-resolution methods} (\textit{e.g.}, SRCNN, SRGAN, and GMFN), DeRefNet demonstrates clear advantages. While traditional methods show limited performance in Sequential CSIST Unmixing, DeRefNet achieves significantly better precision by incorporating multi-frame information and temporal processing. By leveraging both multi-frame data and the dynamic changes in the time sequence, the model is able to capture finer details and temporal relationships, leading to enhanced precision.
This highlights the superiority of DeRefNet in handling dynamic and complex target scenarios. 
Furthermore, DeRefNet benefits from the deep unfolding paradigm and performing super-resolution early in the network rather than at the end. These designs allow DeRefNet to more effectively handle complex target scenarios.

\textbf{3) Compared to deep unfolding methods,} DeRefNet further demonstrates its superiority through enhanced information processing. The use of multi-frame deformable alignment plays a crucial role in this improvement. 
In complex CSIST scenarios, multi-frame deformable alignment captures dynamic inter-frame variations, enabling DeRefNet to surpass other deep unfolding methods. 
This allows DeRefNet to maintain strong performance across various thresholds, showcasing its robustness and ability to perform fine-grained target unmixing in challenging environments.

DeRefNet effectively balances efficiency and performance. With only $0.89M$ learnable parameters, it demonstrates that optimal unmixing performance can be achieved by optimizing a relatively small number of parameters. 
However, its FLOPs reach $15.70 G$, which is relatively higher compared to its lightweight parameter size. 
This increase is due to the use of deformable convolutions, which enhance spatial adaptability and deeper feature extraction. 
Despite its moderate FLOPs, DeRefNet achieves an average processing speed of 367 FPS, which far exceeds the typical acquisition rate of infrared video sequences (usually 25--30 FPS). This indicates strong potential for real-time deployment in practical scenarios.
Furthermore, DeRefNet’s mAP of 51.55 showcases its superior efficiency and exceptional target unmixing capabilities compared to other models with similar parameters or FLOPs.

In conclusion, the DeRefNet model excels in handling Sequential CSIST Unmixing, particularly in high-precision recognition and managing complex scenarios. 
It achieves the best mAP performance among various models, highlighting the superiority of the deep unfolding paradigm and deformable alignment in Sequential CSIST Unmixing. 
These results position DeRefNet as a leading model for similar applications and provide valuable insights for the future development of Sequential CSIST Unmixing, demonstrating significant practical value and broad application potential.

\begin{figure*}[b]
    \centering
    \includegraphics[width=.98\textwidth]{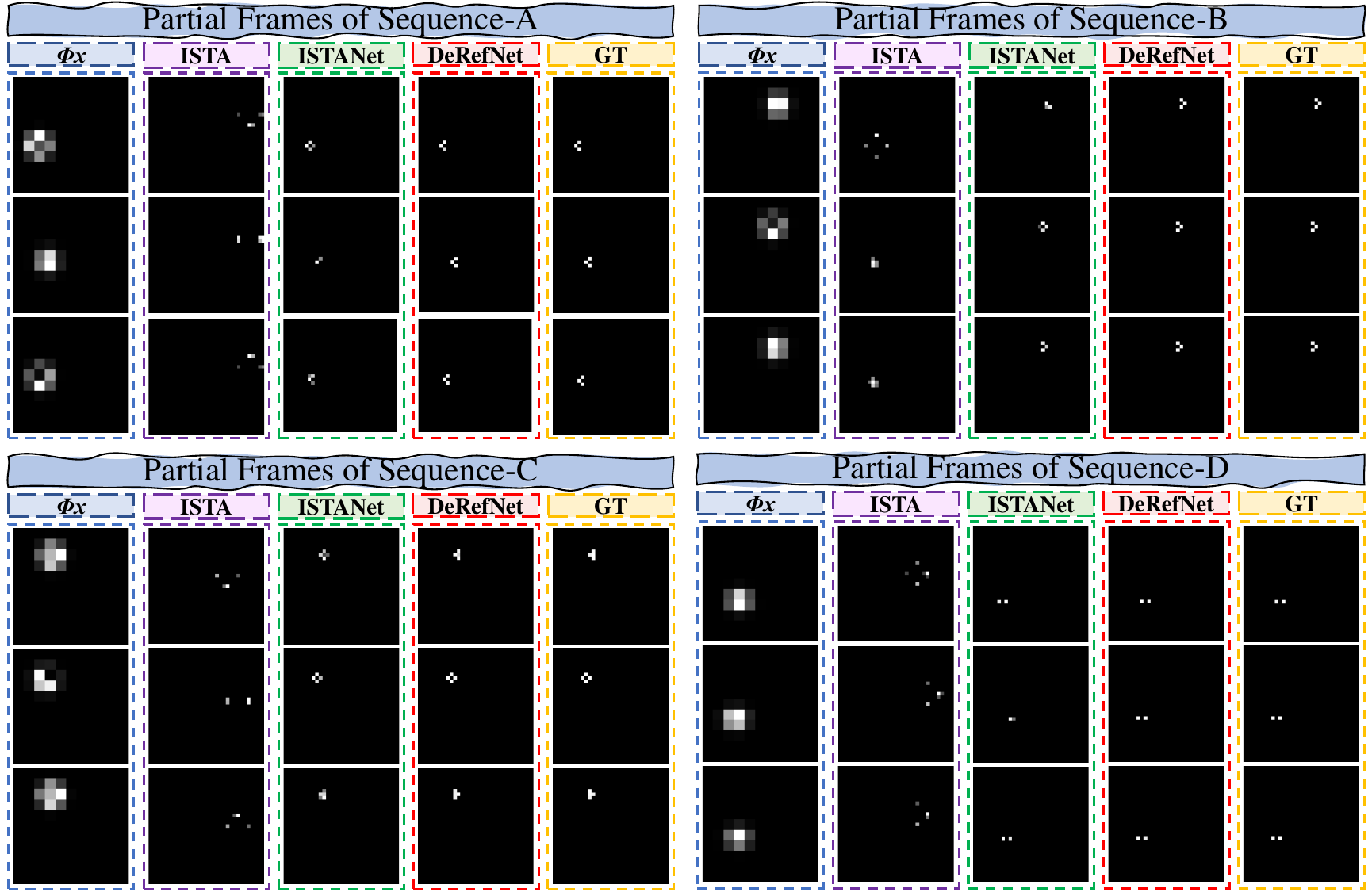}
    \caption{Visual comparison of unmixing performance across different methods on closely-spaced infrared small target sequences. From left to right: input aliased observations ($\Phi$x), traditional iteration-based approach, ISTA, deep unfolding based approach, ISTA-Net, CSIST Unmixing approach, DeRefNet, and ground truth target distribution. DeRefNet demonstrates superior capability in accurately separating individual targets while maintaining clear boundaries, especially in CSIST scenarios where traditional methods struggle with blurred and overlapping features.}
    \label{fig:result}
\end{figure*}

\subsection{Visual Analysis} \label{subsec:visualization}

In order to more intuitively highlight the superiority of our model, we provide the visualizations of some experimental results.

As shown in Table~\ref{table:Comparison with State-of-the-Arts}, while these IRSTD methods achieve high detection rates (some models' Pds even up to 100\%), they fundamentally address a different problem scope. All IRSTD approaches, regardless of their detection accuracy, can only provide bounding box localization and cannot determine the number of individual targets within clustered infrared signatures. This represents a fundamental limitation when dealing with closely-spaced infrared small targets (CSIST), where multiple targets may appear as a single blurred spot due to atmospheric effects, sensor limitations, or proximity.

\begin{table}[H]
  \renewcommand\arraystretch{1.2}
  \footnotesize
  \centering
  \caption{Traditional Detection Methods Performance Overview: Supplementary Evaluation Results for Comparative Visualization Studies}
  \setlength{\tabcolsep}{4pt}
  \begin{tabular}{ccccccc}
  \toprule
  \textbf{MODEL} & \textbf{IoU} ($\%$) & \textbf{Pd} ($\%$) & \textbf{Fa} \\
  \hline
  MSHNet ~\cite{liu2024infrared} & 99.27 & 16.67 & 0.0315 \\
  L2SKNet ~\cite{wu2024saliency} & 99.96 & 100.00 & 0.0001 \\
  IRSAM ~\cite{zhang2024irsam} & 99.67 & 100.00 & 0.0017 \\
  DNANet ~\cite{DNANet} & 99.22 & 99.36 & 0.0004 \\
 \bottomrule
 \label{table:Comparison with State-of-the-Arts}
\end{tabular}
\vspace{-1\baselineskip}
\end{table}

As illustrated in Figure~\ref{fig:comparison}, IRSTD methods treat clustered targets as single entities, providing only binary detection masks. In contrast, our temporal deformable alignment mechanism enables sub-pixel unmixing within each detected region, revealing the actual number of targets and their individual signatures—information that is completely inaccessible to traditional IRSTD approaches. Moreover, we have also examined other representative IRSTD methods, including ISNet~\cite{zhang2022isnet}, RKFormer~\cite{zhang2022rkformer}, IRPruneDet~\cite{zhang2024irprunedet}, and the recent work Unleashing the Power of Generic Segmentation Model~\cite{zhang2024unleashing}. Despite their promising detection performance, these methods are similarly limited to generating binary detection masks and are inherently incapable of revealing the precise number or sub-pixel positions of targets within a densely clustered region. This further underscores the uniqueness and necessity of our sub-pixel unmixing approach for resolving CSIST cases.

As shown in Fig.~\ref{fig:result},
the image presents the unmixing effects for subsequences across multiple sequences. The first column, $\Phi$x, represents closely-spaced infrared small targets images obtained via optical lenses, while the last column, ground truth, shows the actual target distribution. The middle three columns display the unmixing performance of ISTA, ISTA-Net, and DeRefNet.

Across varying numbers of sub-targets, the models show notable differences in unmixing quality.
ISTA struggles to restore the positions, contours, and quantity of aliased small targets, especially when they are densely packed, resulting in blurred and overlapping features, making it difficult to distinguish each sub-target. 
Compared to ISTA, ISTA-Net has a stronger feature extraction capability, with more accurate target localization. This highlights the advantage of model-based deep learning over traditional only model-based iterative algorithms in feature extraction, particularly when sufficient data is available. 
However, as the number of targets increases, ISTA-Net struggles to achieve clear separation, resulting in blurred or overlapping restoration outcomes. Overall, ISTA-Net outperforms ISTA in unmixing, though both fall short of fully separating dense sub-targets.

In contrast, DeRefNet consistently excels, accurately disentangling each sub-target and preserving clarity, even in densely packed scenes. This superior un-aliasing capability highlights our work’s robustness and superior performance across different densities, achieving the most precise target separation results.

%% file: contents/8-analysis.tex




%% file: contents/9-conclusion.tex

\section{Conclusion} \label{sec:conclusion}

In this paper, we propose a novel task, namely Sequential CSIST Unmixing, along with the DeRefNet framework, which consists of three key components: a sparsity-driven feature extraction module, a positional encoding module and a Temporal Deformable Feature Alignment (TDFA) module.
This is the first study to introduce the deep unfolding paradigm into the design of Sequential CSIST Unmixing.
Through comparative studies and extensive experiments, we demonstrate that the proposed DeRefNet framework effectively addresses the CSIST energy aliasing problem in infrared images, achieving unmixing and sub-pixel localization.
Additionally, we have constructed an open-source ecosystem for infrared target unmixing, which includes sequential benchmark dataset and a toolkit, providing valuable resources for related research. 

\section*{Acknowledgment}

The authors would like to thank the editor and the anonymous reviewers for their critical and constructive comments and suggestions.
We acknowledge the Tianjin Key Laboratory of Visual Computing and Intelligent Perception (VCIP) for their essential resources. Computation is partially supported by the Supercomputing Center of Nankai University (NKSC).

%% file: contents/10-biography.tex

\begin{IEEEbiography}[{\includegraphics[width=1in,height=1.25in,clip,keepaspectratio]{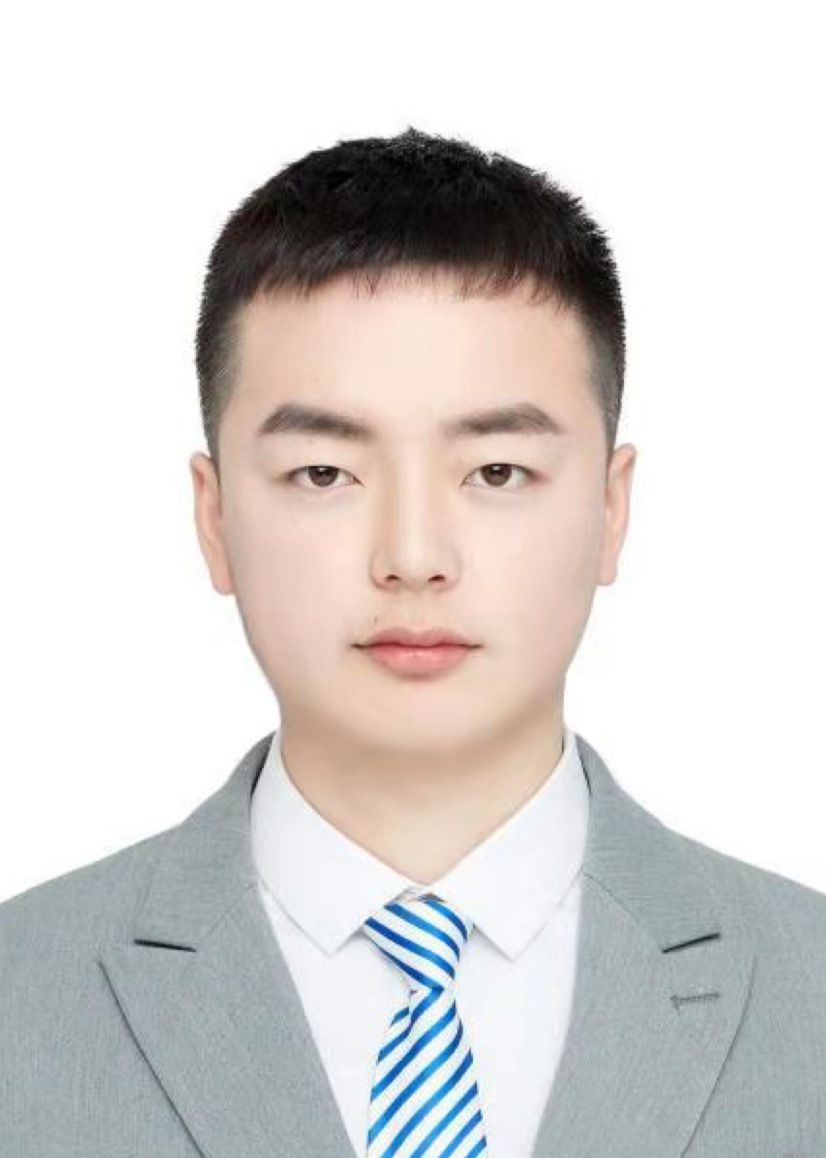}}]{Ximeng Zhai} is currently pursuing a Master's degree at the Xi'an Institute of Optics and Precision Mechanics (XIOPM), University of Chinese Academy of Sciences, China. He received his Bachelor's degree from Tianjin University. His research interests encompass infrared image processing and object detection.
\end{IEEEbiography}

\begin{IEEEbiography}[{\includegraphics[width=1in,height=1.25in,clip,keepaspectratio]{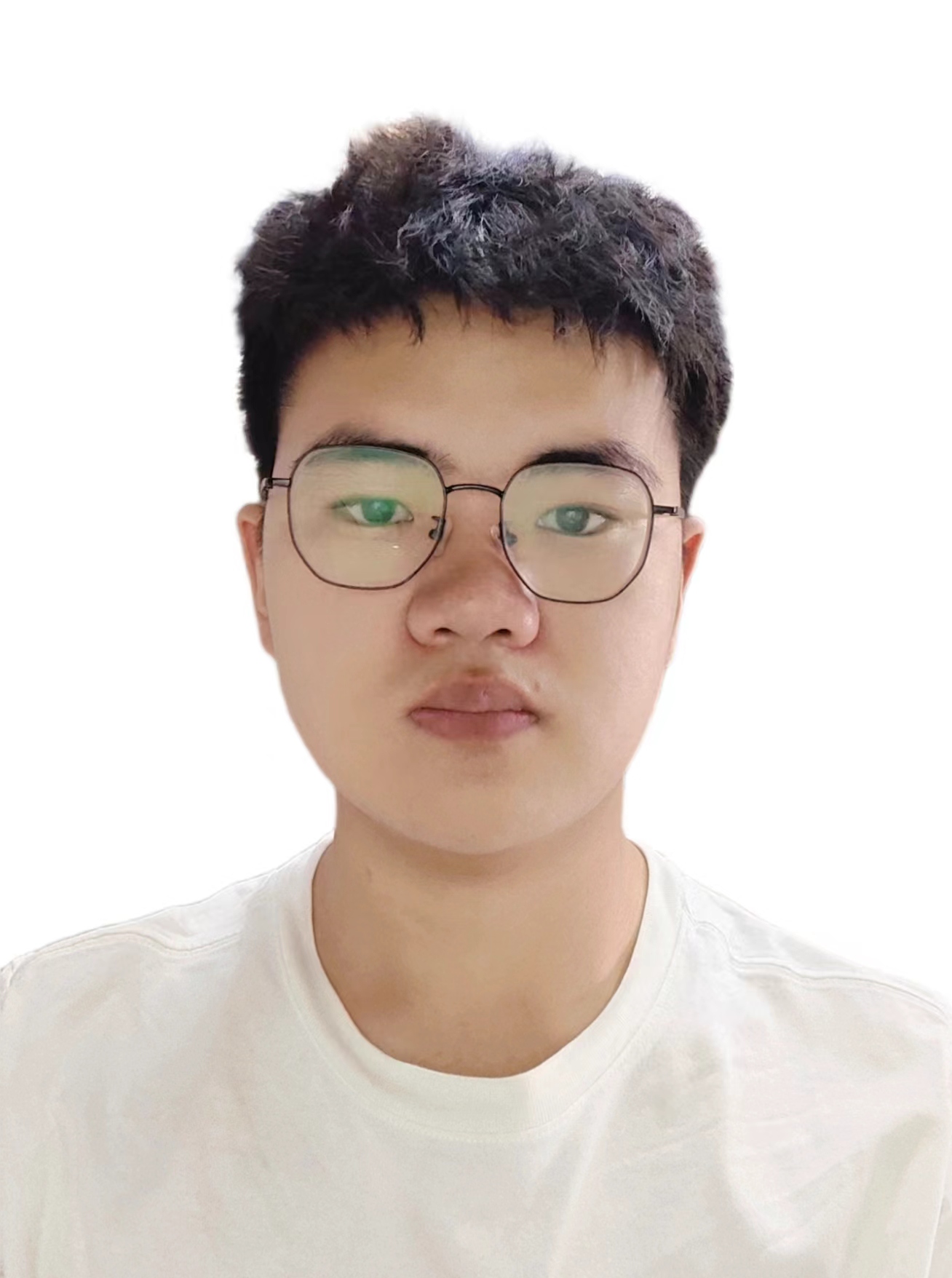}}]{Bohan Xu} is currently pursuing his Bachelor's degree in Electronic Information Engineering at the School of Information Science and Engineering, Henan University of Technology. His primary research focus is on infrared small target super-resolution.
\end{IEEEbiography}

\begin{IEEEbiography}[{\includegraphics[width=1.1in,height=1.35in,clip,keepaspectratio]{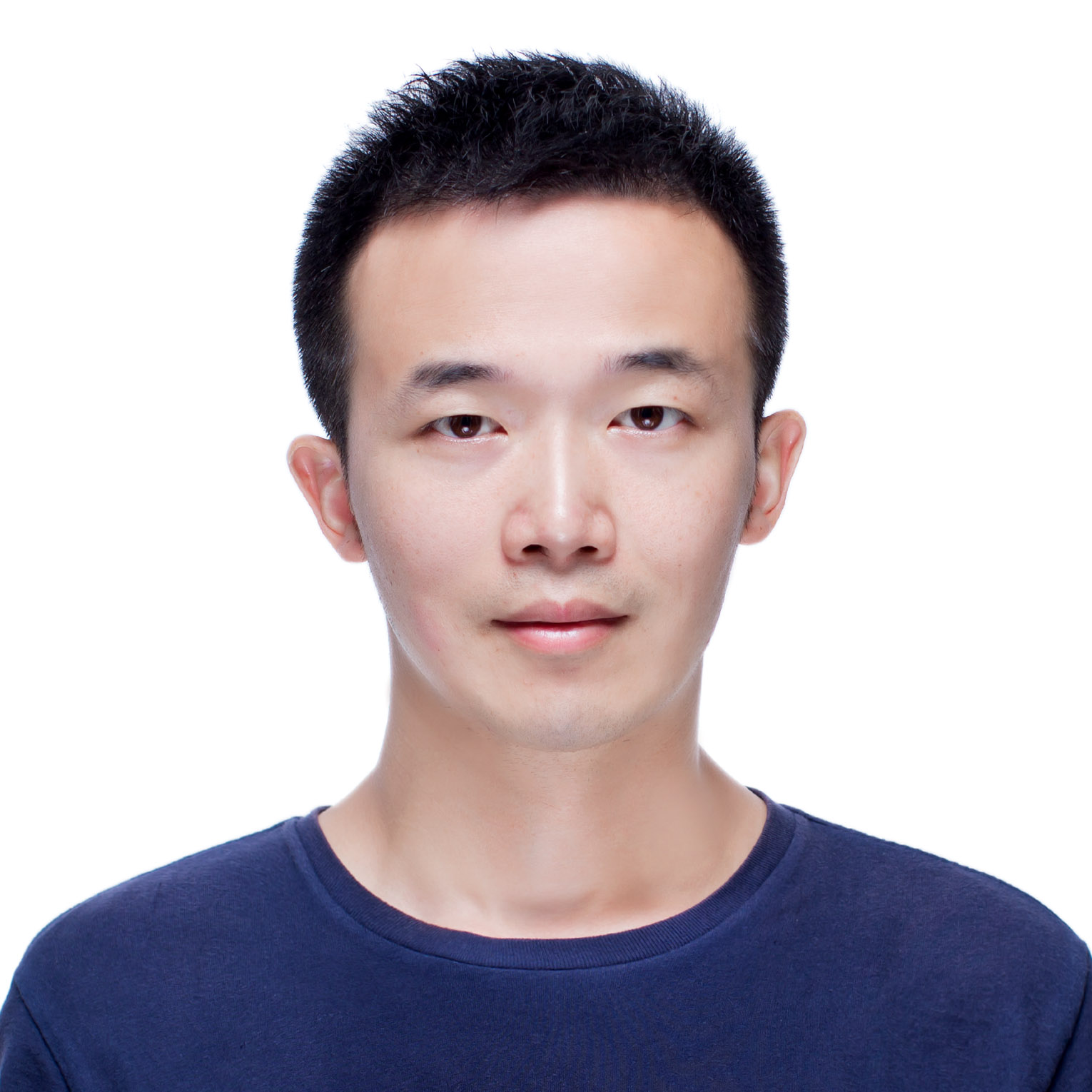}}]{Yaohong Chen} received his M.S. and Ph.D. degree from Xi'an Jiaotong University and University of Chinese Academy of Sciences in 2013 and 2022, respectively. He was a visiting scholar in the Department of Electrical and Computer Engineering of Johns Hopkins University during the 2019 to 2020. He is currently an associate research fellow with the Xi'an Institute of Optics and Precision Mechanics, Chinese Academy of Sciences. His research interests include the infrared imaging systems, infrared image processing and infrared imaging circuit.
\end{IEEEbiography}

\begin{IEEEbiography}[{\includegraphics[width=1.1in,height=1.35in,clip,keepaspectratio]{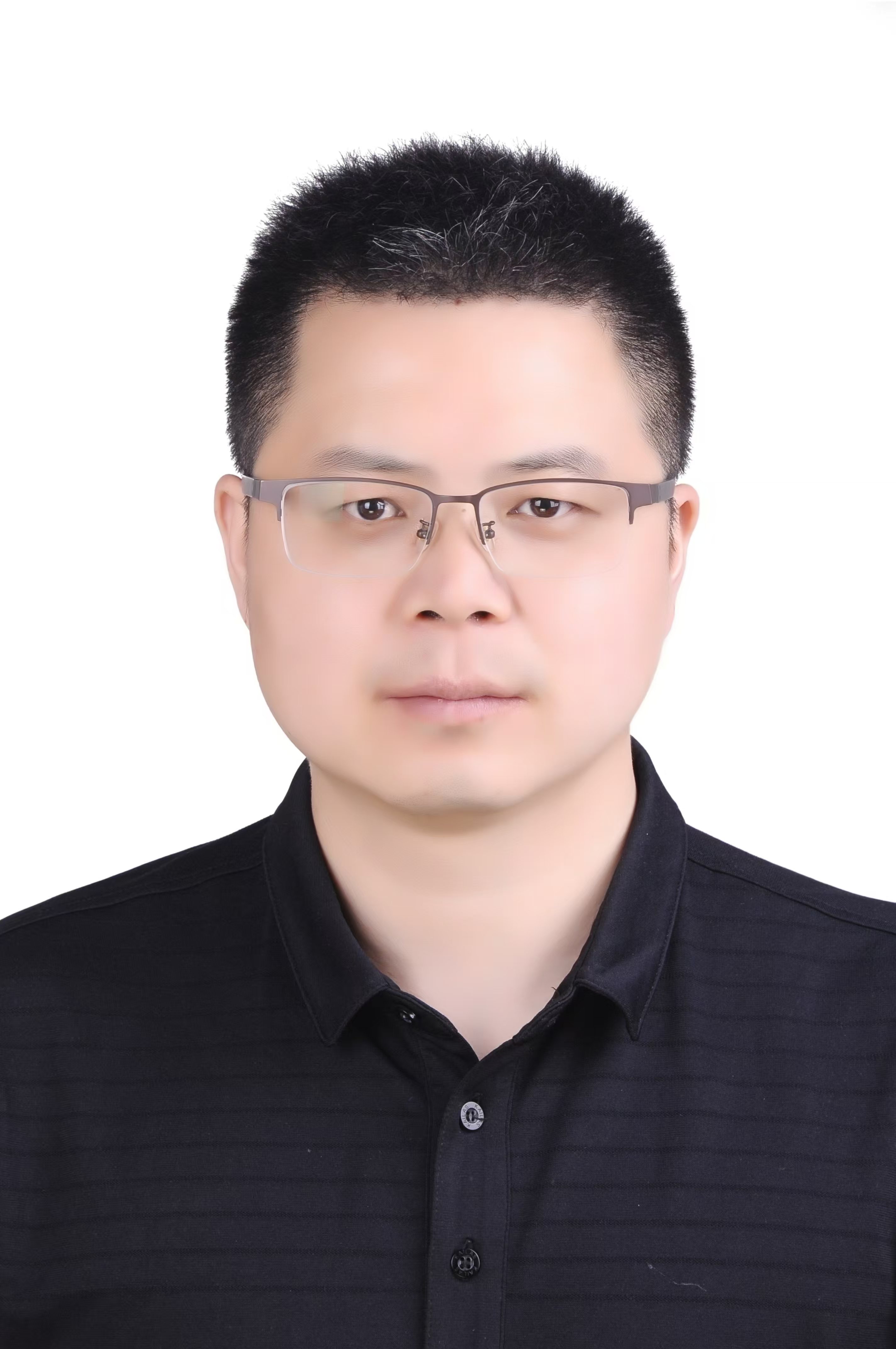}}]{Hao Wang} received his M.S. and Ph.D. degrees from the University of Electronic Science and Technology and the University of Chinese Academy of Sciences in 2008 and 2017, respectively. He is currently a research fellow at the Xi'an Institute of Optics and Precision Mechanics, Chinese Academy of Sciences. His research focuses on aerospace visual imaging and image processing.
\end{IEEEbiography}

\begin{IEEEbiography}[{
\includegraphics[width=1.45in,height=1.3in,clip,keepaspectratio]{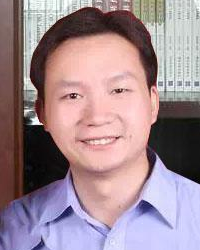}}]{Kehua Guo} (M'13) received the Ph.D. degree in Computer Science and Technology from Nanjing University of Science and Technology, Nanjing, China, in 2008. He is currently a professor with Central South University, Changsha, China. He has served as a Guest Editor, Workshop Chair, Publicity Chair, Technical Program Committee Member, and Reviewer of international journals/conference proceedings. His research interests include artificial intelligence, big data, and image processing.
\end{IEEEbiography}

\begin{IEEEbiography}[{\includegraphics[width=1in,height=1.25in,clip,keepaspectratio]{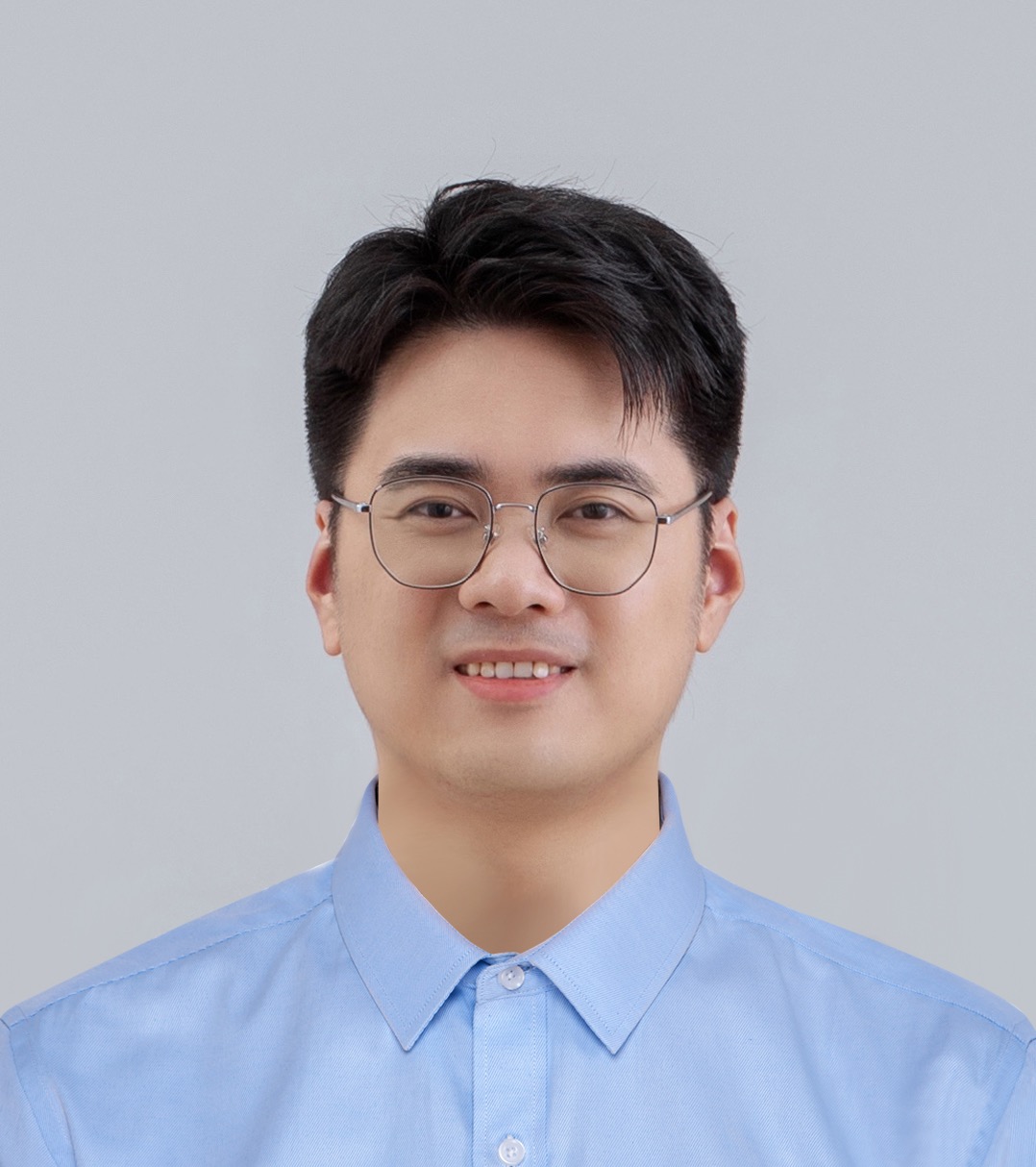}}]{Yimian Dai} 
(Member, IEEE) received the B.E. degree in information engineering and the Ph.D. degree in signal and information processing from Nanjing University of Aeronautics and Astronautics, Nanjing, China, in 2013 and 2020, respectively.
From 2021 to 2024, he was a Postdoctoral Researcher with the School of Computer Science and Engineering, Nanjing University of Science and Technology, Nanjing, China. 
He is currently an Associate Professor with the College of Computer Science, Nankai University, Tianjin, China.
His research interests include computer vision, deep learning, and their applications in remote sensing.
For more information, please visit the link (\href{https://yimian.grokcv.ai/}{https://yimian.grokcv.ai/}).
\end{IEEEbiography}